\documentclass[romanappendices,journal,compsoc]{IEEEtran}
\usepackage{graphicx}
\usepackage{amsmath}
\usepackage{amssymb}
\usepackage{booktabs}
\usepackage[ruled,vlined,linesnumbered,resetcount]{algorithm2e}
\usepackage{multirow}
\usepackage{varwidth}
\usepackage{textcomp}
\usepackage{subcaption}
\usepackage[pagebackref,breaklinks,colorlinks]{hyperref}
\hypersetup{
    unicode=false,          % non-Latin characters in Acrobat’s bookmarks
    pdftoolbar=true,        % show Acrobat’s toolbar?
    pdfmenubar=true,        % show Acrobat’s menu?
    pdffitwindow=false,     % window fit to page when opened
    pdfstartview={FitH},    % fits the width of the page to the window
    colorlinks=true,       % false: boxed links; true: colored links
    linkcolor=red,          % color of internal links (change box color with linkbordercolor)
    citecolor=green,        % color of links to bibliography
    filecolor=cyan,         % color of file links
    urlcolor=magenta        % color of external links
}
\usepackage{ragged2e}

\graphicspath{{Figures/}}

\makeatletter
\newcommand{\nosemic}{\renewcommand{\@endalgocfline}{\relax}}% Drop semi-colon ;
\newcommand{\dosemic}{\renewcommand{\@endalgocfline}{\algocf@endline}}% Reinstate semi-colon ;
\let\oldnl\nl% Store \nl in \oldnl
\newcommand{\nonl}{\renewcommand{\nl}{\let\nl\oldnl}}% Remove line number for one line
\makeatother

\begin{document}

\title{ADVISE: {AD}aptive Feature Relevance and {VI}Sual {E}xplanations for Convolutional Neural Networks}

\author{Mohammad Mahdi~Dehshibi, Mona~Ashtari-Majlan, Gereziher~Adhane,
        David~Masip,~\IEEEmembership{Senior~Member,~IEEE}% <-this % stops a space
\IEEEcompsocitemizethanks{\IEEEcompsocthanksitem Department of Computer Science, Universitat Oberta de Catalunya, 08018, Barcelona, Spain.\protect\\
Corresponding author: M. M. Dehshibi (e-mail: mohammad.dehshibi@yahoo.com).}% <-this % stops an unwanted space
}

\IEEEtitleabstractindextext{
\justify
\begin{abstract}
To equip Convolutional Neural Networks (CNNs) with explainability, it is essential to interpret how opaque models take specific decisions, understand what causes the errors, improve the architecture design, and identify unethical biases in the classifiers. This paper introduces ADVISE, a new explainability method that quantifies and leverages the relevance of each unit of the feature map to provide better visual explanations. To this end, we propose using adaptive bandwidth kernel density estimation to assign a relevance score to each unit of the feature map with respect to the predicted class. We also propose an evaluation protocol to quantitatively assess the visual explainability of CNN models. We extensively evaluate our idea in the image classification task using AlexNet, VGG16, ResNet50, and Xception pretrained on ImageNet. We compare ADVISE with the state-of-the-art visual explainable methods and show that the proposed method outperforms competing approaches in quantifying feature-relevance and visual explainability while maintaining competitive time complexity. Our experiments further show that ADVISE fulfils the sensitivity and implementation independence axioms while passing the sanity checks. The implementation is accessible for reproducibility purposes on \url{https://github.com/dehshibi/ADVISE}.
\end{abstract}

}

\maketitle
\IEEEraisesectionheading{\section{Introduction}\label{sec:intro}}
\IEEEPARstart{C}{onvolutional} Neural Networks (CNNs) have gained significant prominence with the potential to outperform expectations in various computer vision tasks such as image classification~\cite{chen2019all,adhane2021deep,adhane2021use}, object detection~\cite{tan2020efficientdet}, semantic segmentation~\cite{mehta2018espnet}, image captioning~\cite{chen2017sca}, and human behaviour analysis~\cite{dehshibi2021deep}. However, this sub-symbolism (also known as the opaque or black-box model) is vulnerable to the underlying barrier of \textit{explainability} in response to critical questions like how a particular trained model arrives at a decision, how certain it is about its decision, if and when it can be trusted, why it makes certain mistakes, and in which part of the learning algorithm or parametric space correction should take place~\cite{lipton2018mythos,arrieta2020explainable}. Explainability in CNNs is linked to post-hoc explainability~\cite{guidotti2018survey} and, as proposed by Arrieta et al.~\cite{arrieta2020explainable}, relies on model simplification~\cite{zhang2019interpreting,ribeiro2016should,kim2021lightweight}, feature-relevance estimation~\cite{binder2016layer,nguyen2016synthesizing,lundberg2017unified,samek2017explainable}, visualisation~\cite{zeiler2014visualizing,mahendran2015understanding,li2016convergent,selvaraju2017grad,taha2020generic,jiang2021layercam}, and architectural modification~\cite{lin2014network,donahue2017long,seo2017interpretable} to convert a non-interpretable model into an explainable one.

While model simplification and architectural modification techniques have been used to make CNNs interpretable, their associated complexity grows as the number of layers and parameters increases. Furthermore, several  studies~\cite{bau2017network,olah2017feature,arrieta2020explainable} have shown that altering CNNs may result in the spontaneous appearance of a disentangled representation~\cite{gonzalez2018semantic,zhou2015object}, which is not only unrelated to the model's initial intention but also challenging to interpret. As a result, the emphasis in explaining CNNs has shifted toward feature-relevance and visualisation methods.

\begin{figure*}[!htp]
  \centering
  \begin{subfigure}{0.15\linewidth}
    \includegraphics[width=1\linewidth]{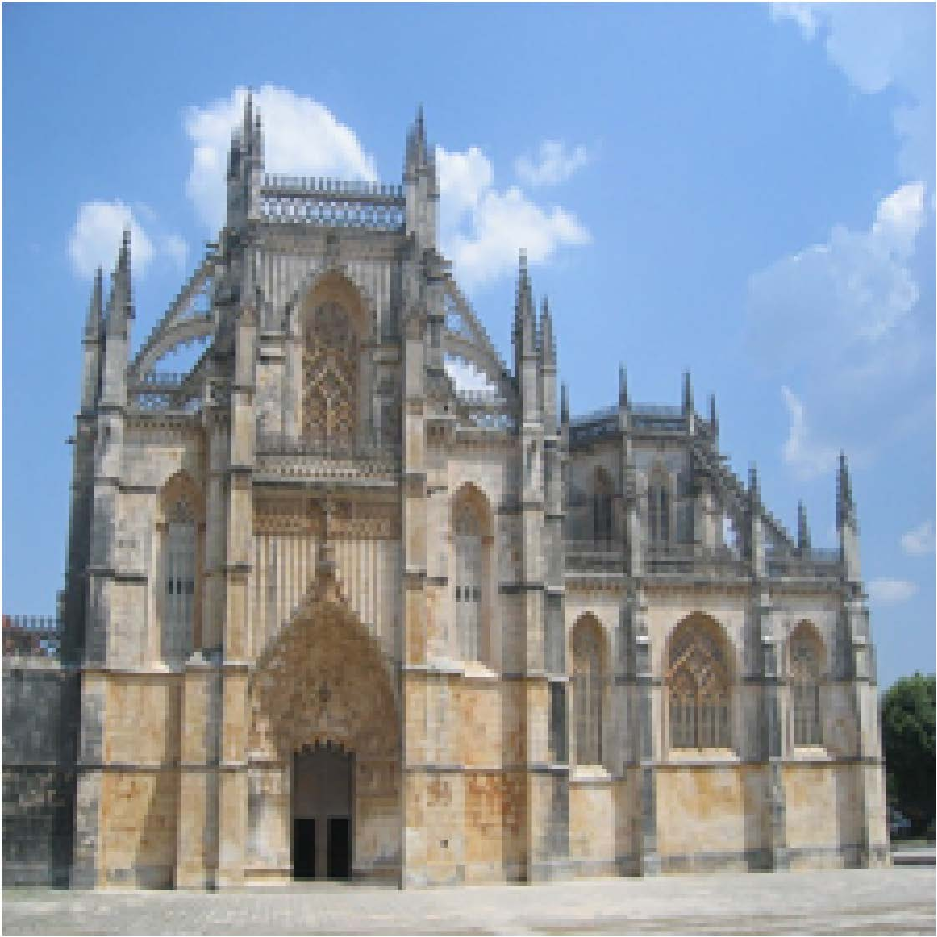}
    \caption{}
    \label{fig:01a}
  \end{subfigure}
  \hfill
  \begin{subfigure}{0.17\linewidth}
    \includegraphics[width=1\linewidth]{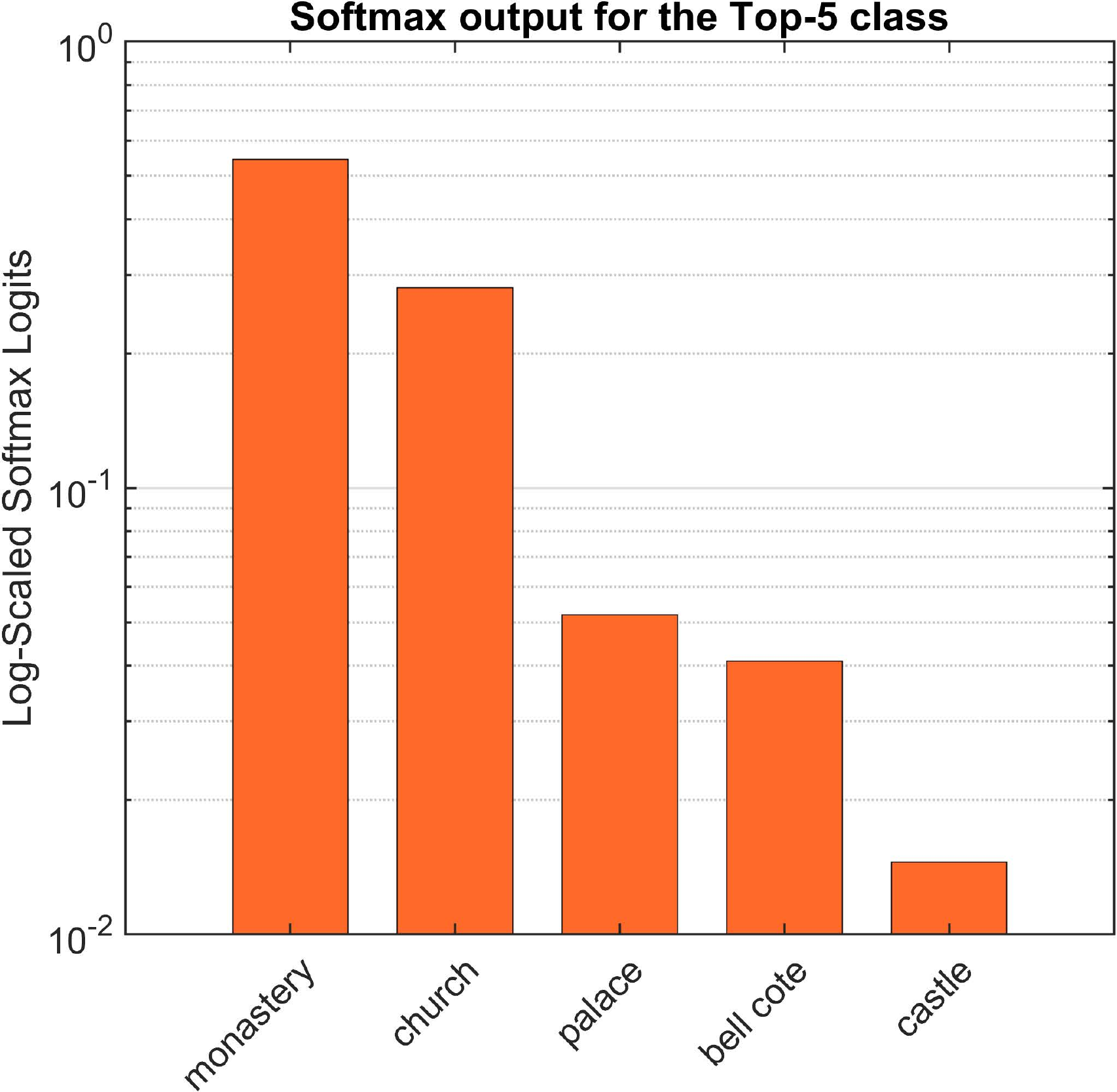}
    \caption{}
    \label{fig:01_vgg_b}
  \end{subfigure}
  \hfill
  \begin{subfigure}{0.15\linewidth}
    \includegraphics[width=1\linewidth]{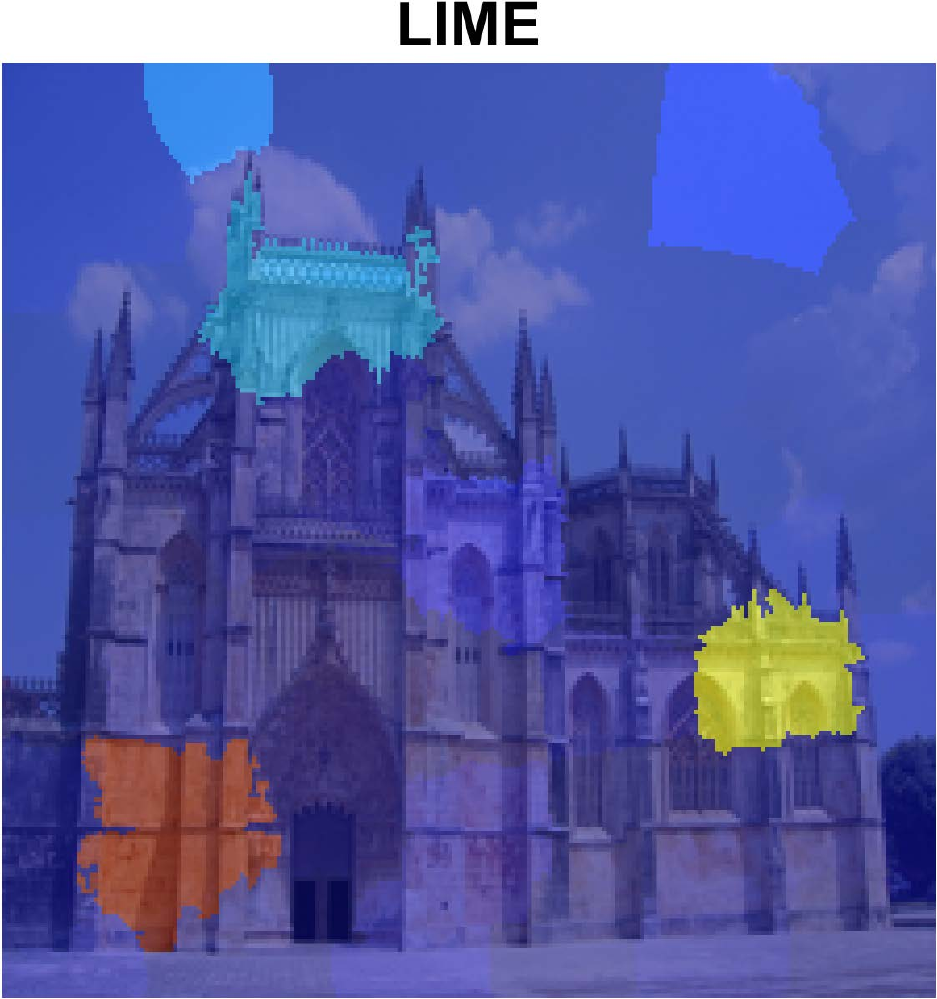}
    \caption{}
    \label{fig:01_vgg_c}
  \end{subfigure}
  \hfill
  \begin{subfigure}{0.15\linewidth}
    \includegraphics[width=1\linewidth]{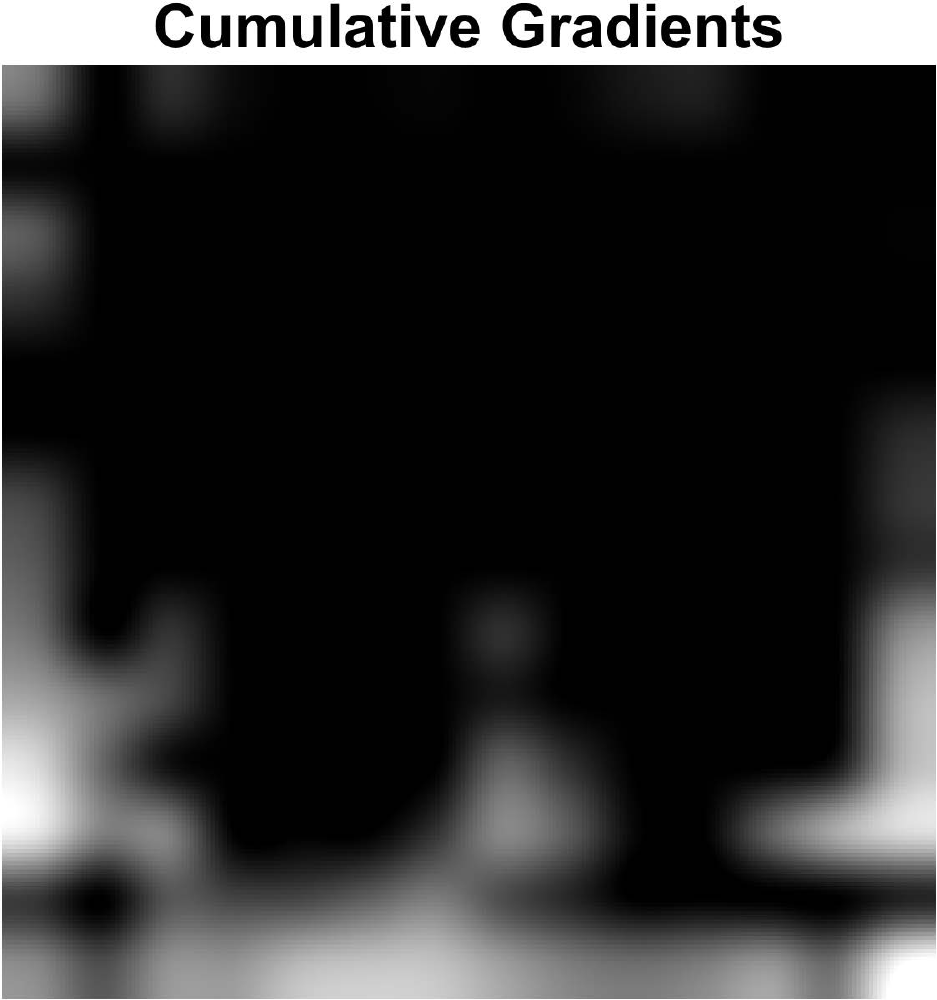}
    \caption{}
    \label{fig:01_vgg_d}
  \end{subfigure}
  \hfill
  \begin{subfigure}{0.17\linewidth}
    \includegraphics[width=1\linewidth]{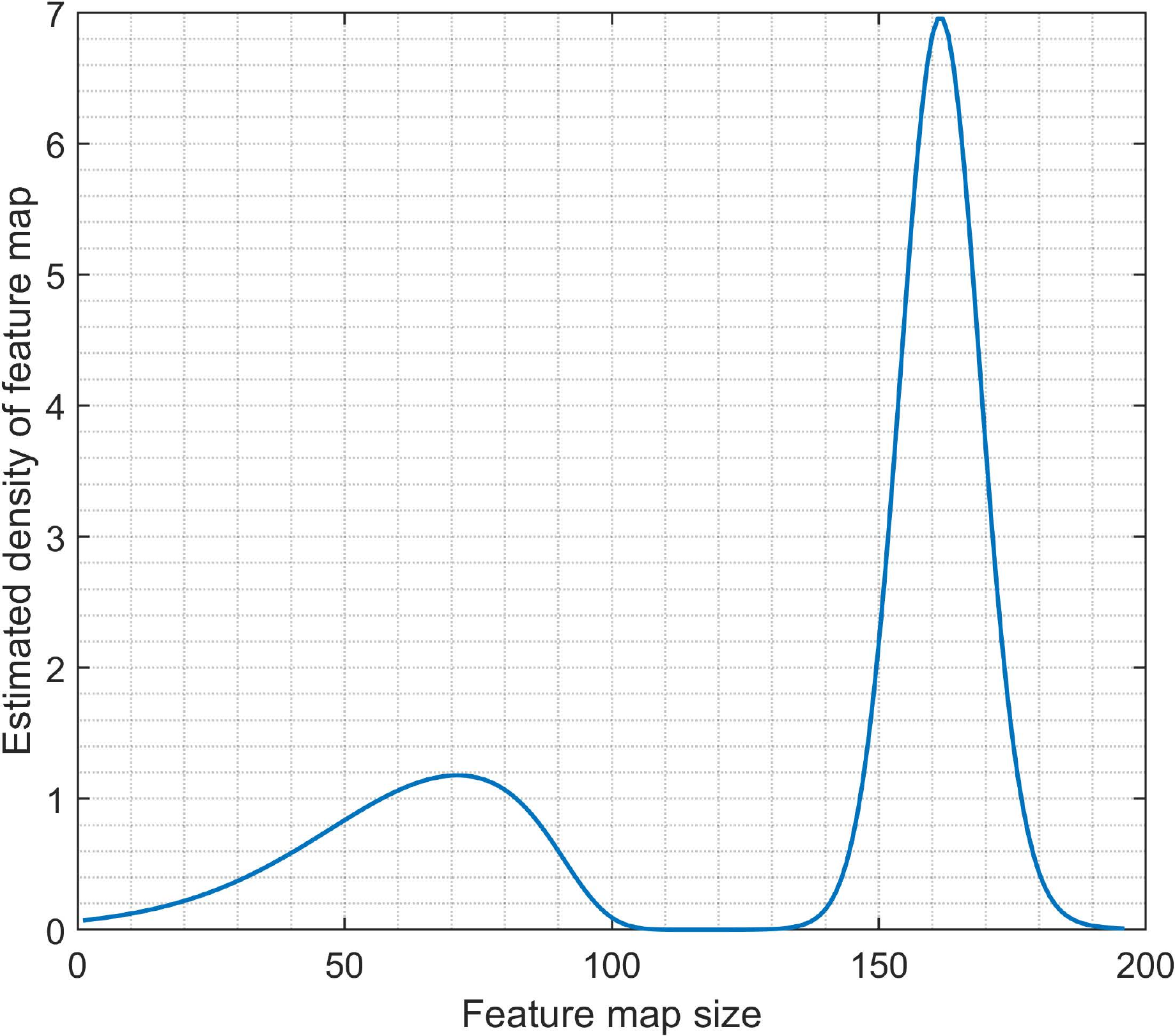}
    \caption{}
    \label{fig:01_vgg_e}
  \end{subfigure}
  \hfill
  \begin{subfigure}{0.15\linewidth}
    \includegraphics[width=1\linewidth]{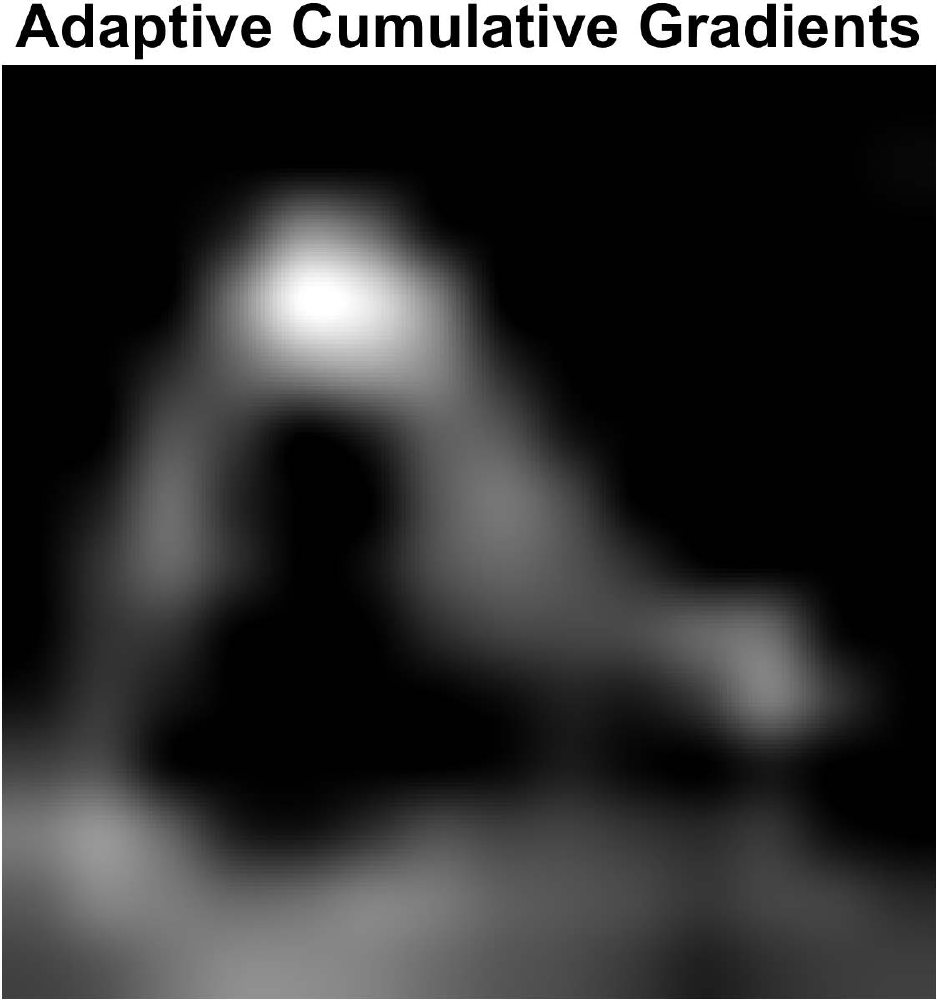}
    \caption{}
    \label{fig:01_vgg_f}
  \end{subfigure}
  \hfill
  \begin{subfigure}{0.15\linewidth}
    \includegraphics[width=1\linewidth]{Fig01a}
    \caption{}
  \end{subfigure}
  \hfill
  \begin{subfigure}{0.17\linewidth}
    \includegraphics[width=1\linewidth]{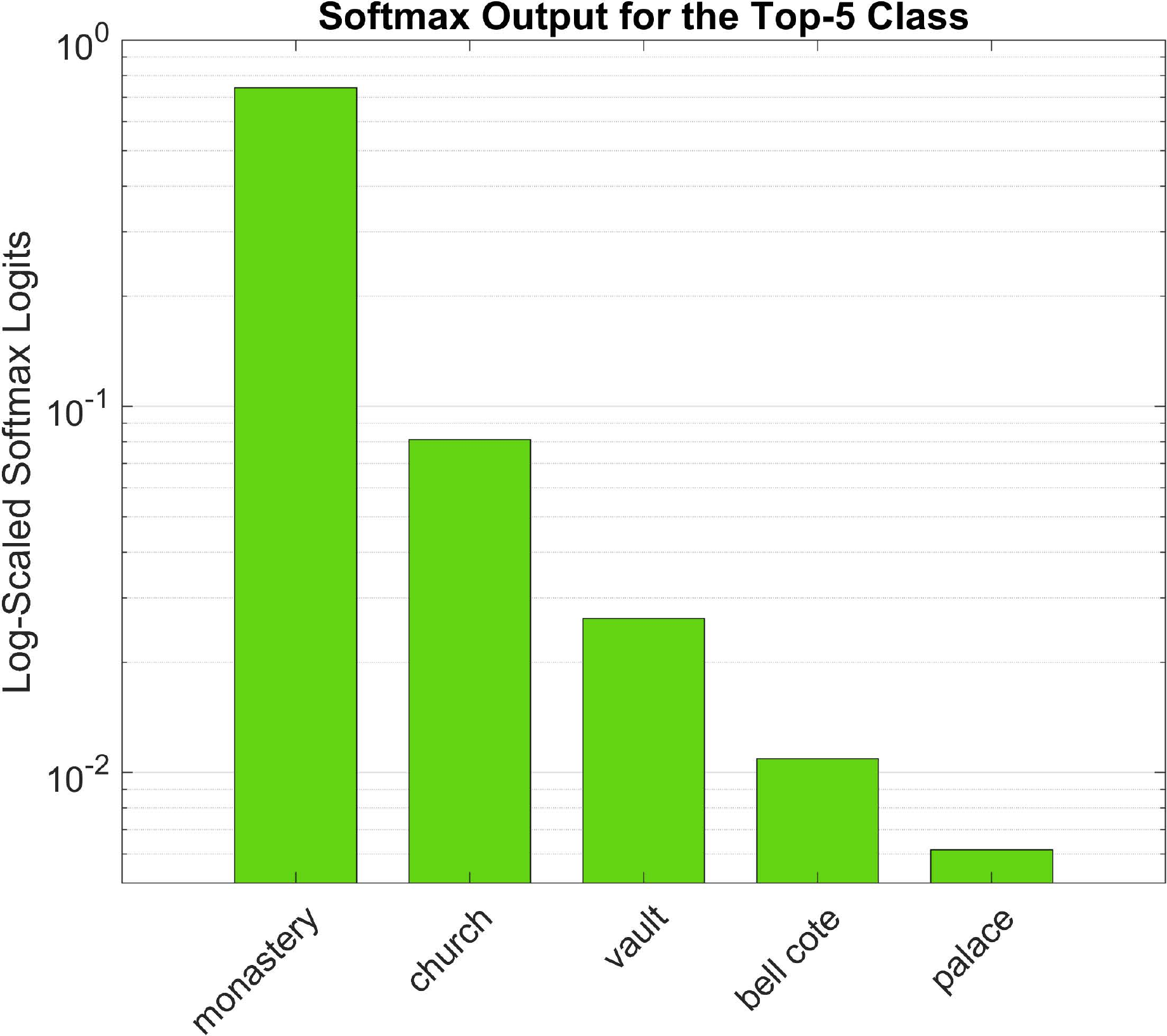}
    \caption{}
    \label{fig:01_xception_b}
  \end{subfigure}
  \hfill
  \begin{subfigure}{0.15\linewidth}
    \includegraphics[width=1\linewidth]{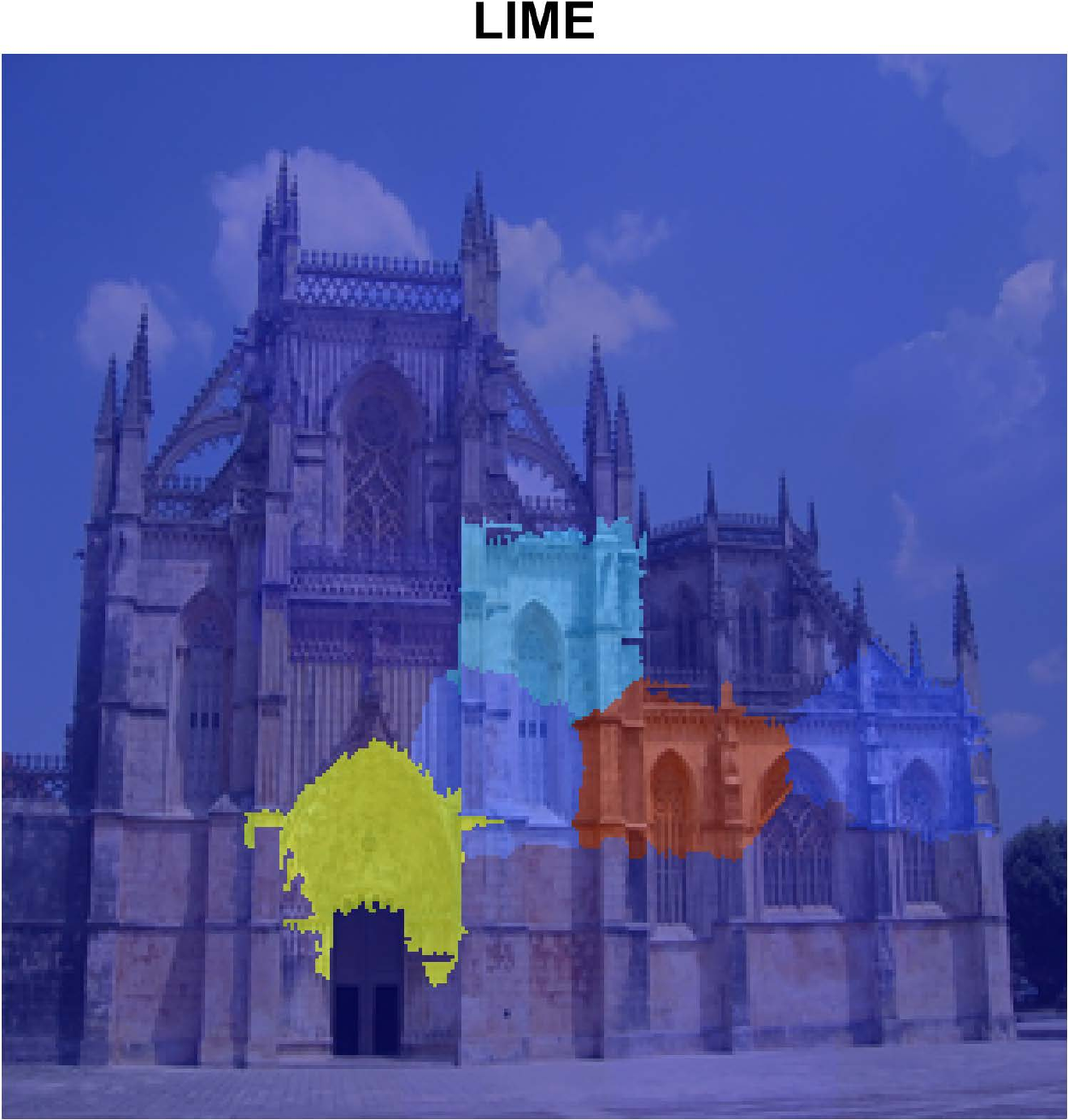}
    \caption{}
    \label{fig:01_xception_c}
  \end{subfigure}
  \hfill
  \begin{subfigure}{0.15\linewidth}
    \includegraphics[width=1\linewidth]{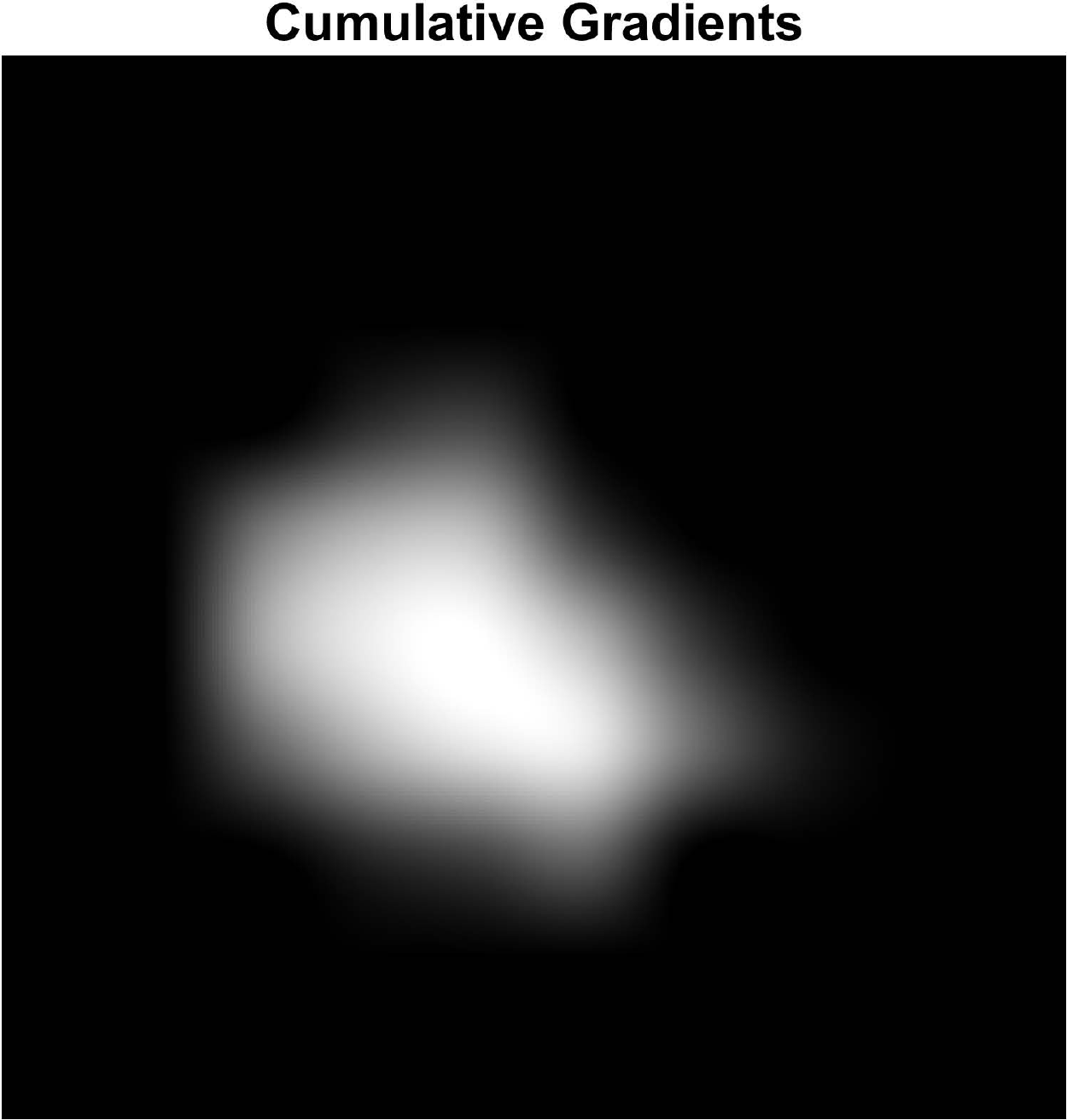}
    \caption{}
    \label{fig:01_xception_d}
  \end{subfigure}
  \hfill
  \begin{subfigure}{0.17\linewidth}
    \includegraphics[width=1\linewidth]{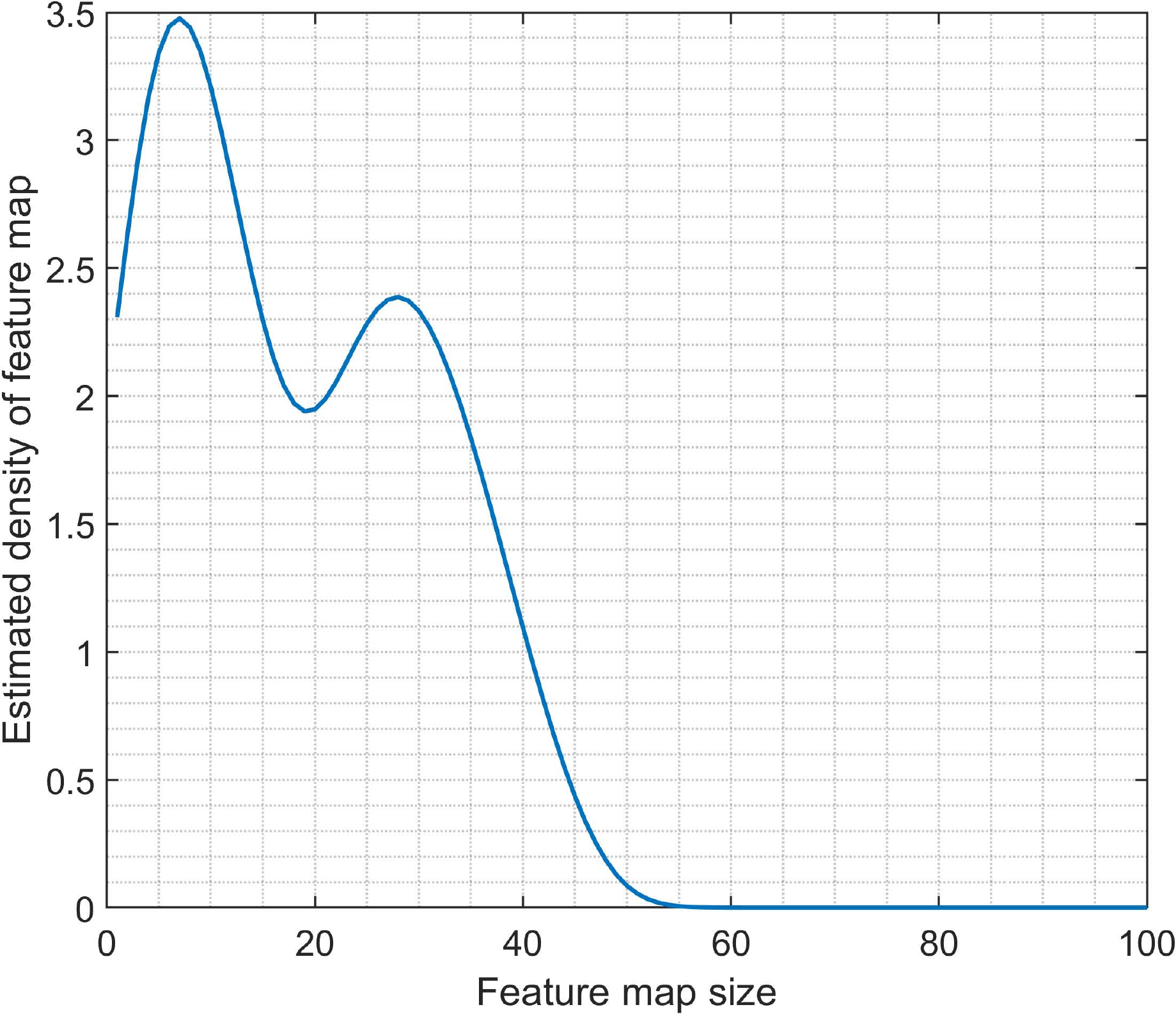}
    \caption{}
    \label{fig:01_xception_e}
  \end{subfigure}
  \hfill
  \begin{subfigure}{0.15\linewidth}
    \includegraphics[width=1\linewidth]{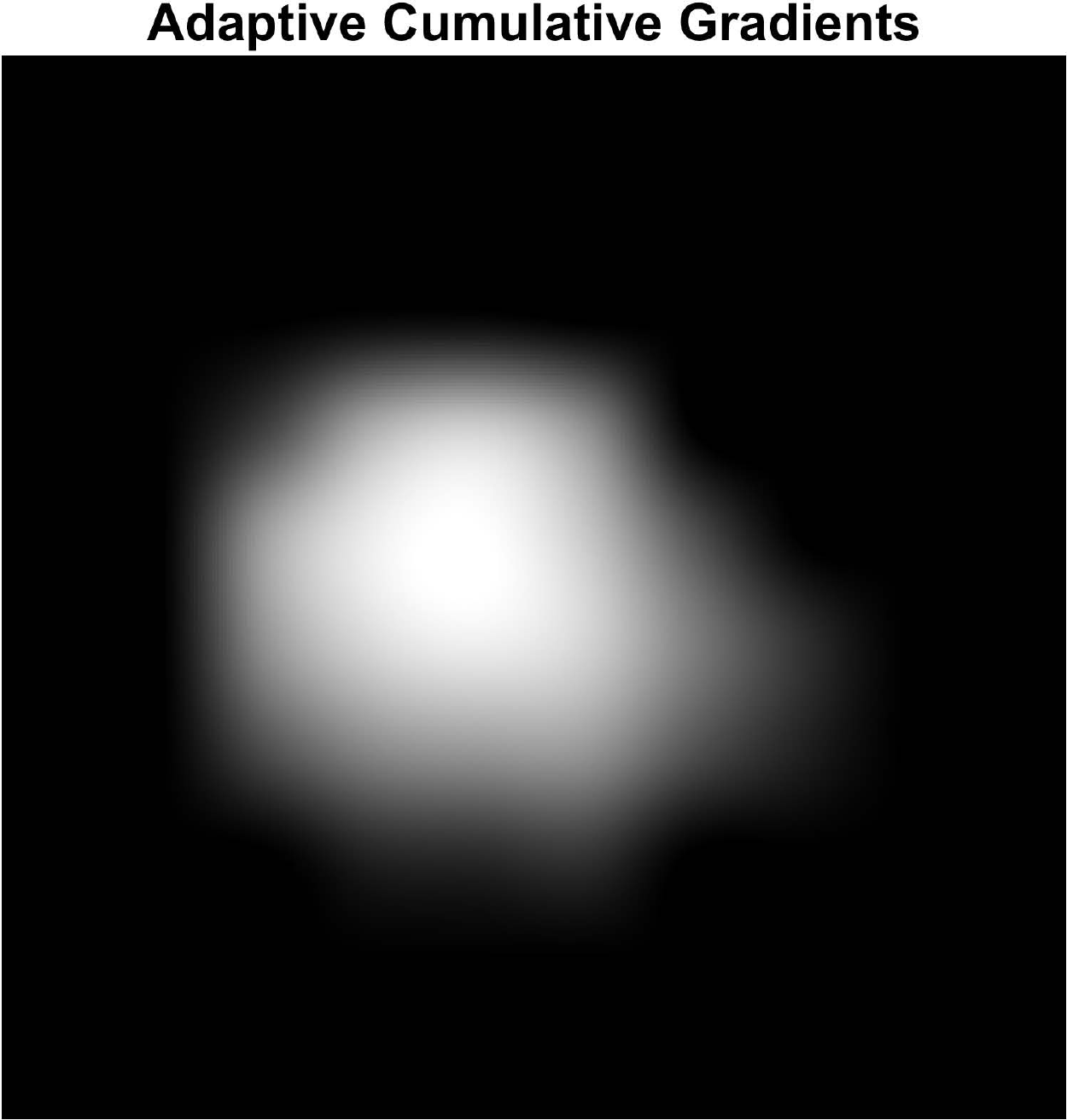}
    \caption{}
    \label{fig:01_xception_f}
  \end{subfigure}
  \caption{The visual and feature-relevance explanations for VGG16~\cite{simonyan2015very} (the first row) and Xception~\cite{chollet2017xception} (the second row) pretrained models on ILSVRC~\cite{russakovsky2015imagenet}. (a, g) The input image. (b, h) The output of the Log-scaled softmax Logits for the top-5 predicted classes. (c, i) Local explanations for the prediction of the input image based on LIME~\cite{ribeiro2016should}. (d, j) Cumulative gradients of the last convolutional layer, where the feature map is scaled up to the resolution of the input image using bilinear interpolation. (e, k) Estimated density of the $k^{\text{th}}$ unit in feature map, which represents 2 peaks. (f, l) Adaptive cumulative gradients of units with 2 peaks in their estimated density.}
  \label{fig:01}
\end{figure*}

Feature visualisation has received much attention because human cognitive skills favour the understanding of visual data. However, feature visualisation methods do not necessarily provide a comprehensive level of explainability and interpretability. For instance, in Figure~\ref{fig:01a}, an identical image is fed into the VGG16 and Xception models, both of which outperform humans on ImageNet classification. Although both models have the exact top-1 prediction with one difference in top-5 prediction, the visual explanations are significantly different (see LIME and Cumulative Gradients in Figure~\ref{fig:01}) and cannot provide users with comprehensive information about how the models made the final decision. Therefore, several studies~\cite{binder2016layer,shrikumar2017learning,lundberg2017unified} focused on feature-relevance approaches, which provide an importance score to each feature for a specific input. However, the visual and feature-relevance explanations are not mutually exclusive when a feature-relevance method can be visualised as a saliency map~\cite{ribeiro2016should}, and a saliency map generated using class activation maps~\cite{zeiler2014visualizing,selvaraju2017grad,jiang2021layercam} can assign importance scores to each pixel.

%%PLACEMENT

In this paper, we propose a method for quantifying the feature-relevance and visualising the latent representations in CNNs. We revisit the relationships between feature maps\footnote{The terms \textit{feature map} and \textit{activation map} are used interchangeable here since the former refers to a mapping of where a specific type of feature can be found in an image, and the latter is a mapping that relates to the activation of different areas of the image.} and their associated gradients by introducing {AD}aptive {VI}Sual {EX}planation (ADVISE). \textbf{ADVISE} estimates the kernel density of gradients with an adaptive bandwidth for each unit in the feature map (see Figures~\ref{fig:01_vgg_e} and~\ref{fig:01_xception_e}) to assign an importance score to each unit. Then, we calculate the cumulative gradient of units with the same importance score for the class of interest to visualise the feature map. In this way, we simultaneously quantify the relevance of each unit and highlight how much the cumulative gradient of units influence the model's decision using the generated saliency map(s) (see Figures~\ref{fig:01_vgg_f} and~\ref{fig:01_xception_f}). We use the proposed method to demonstrate that individual units are significantly more interpretable than cumulative linear combinations of gradient's units.

Our experiment is centred on the image classification task since it allows us to visualise adaptive cumulative gradient attributions and compare ADVISE with attention approaches that focus on global information. We use AlexNet~\cite{krizhevsky2012imagenet}, VGG16~\cite{simonyan2015very}, ResNet50~\cite{he2016deep}, and Xception~\cite{chollet2017xception}, which were trained on the ILSVRC~\cite{russakovsky2015imagenet} in order to decide to which of 1000 classes each image belongs. However, unlike previous approaches, estimating the kernel density of gradients with the adaptive bandwidth can be applied to a wide range of deep learning models without requiring architectural changes or retraining.

The rest of this paper is organised as follows: Section~\ref{sec:survey} surveys the previous studies. The proposed method is detailed in Section~\ref{sec:methodology}. Section~\ref{sec:experiment} presents experimental results. Finally, Section~\ref{sec:conclusion} concludes the paper.

%------------------------------------------------------------------------
\section{Literature Review}
\label{sec:survey}
 
As previously stated, explaining a model by the visualisation (i.e.,, explicit explainability) and feature-relevance (i.e.,, implicit explainability) are not mutually exclusive. In fact, visualisation techniques present complementary ways of visualising the output of feature relevance techniques to aid model interpretation.

In this context, the methods proposed to explain what CNNs learn can be categorised into three  broad categories: (1) those that rely on attention methods by generating class activation maps to interpret how the intermediate layers perceive the external world with respect to the target class without restricting the method to any specific input; (2) those that interpret the decision process using a top-down back-propagation strategy in which the output is mapped back in the input space to determine which parts of the input are discriminative for the output; (3) those that integrate importance over the attribution path and open up the axiomatic \textit{sensitivity} and \textit{implementation invariance} attributions for deep neural networks. These methods are amenable to intriguing visualisations and serve as a basis for discussing missingness in the feature space.

The main idea behind class activation mapping is to achieve class-specific importance for each location of an image by multiplying each feature map by its weight and performing a sum of all weighted feature map values at that location across all channels (units). Following this procedure, a ReLU operation is used to filter out negative activations. The method of calculating the weight for each feature map differs between different attention methods. CAM~\cite{zhou2016learning} obtained the weights from a single fully connected layer that produces the predictions, in which global average pooling is applied to the final convolutional feature maps. Grad-CAM~\cite{selvaraju2017grad} improved on CAM by applying class-specific gradients to each feature map at each location and averaging the gradients of each feature map unit as its weight. Grad CAM++~\cite{chattopadhay2018grad} generates a visual explanation for the corresponding class label by using a weighted combination of the positive partial derivatives of the last convolutional layer feature maps with respect to a specific class score as weights. Score-CAM~\cite{wang2020score} eliminates gradient dependence by masking the input image with the activation map generated with respect to the target class at different network layers and passing it through the network to obtain the prediction score. Finally, the weight for each feature map is calculated by the normalised sum of the obtained scores. Layer-CAM~\cite{jiang2021layercam} utilises the backward class-specific gradients, in which the gradient with respect to the class of interest is calculated for each unit in the feature map, and the units with positive gradient values are used as weights.

Zhang et al.~\cite{zhang2018top} introduced a top-down back-propagation approach to compute neuron significance towards a model that passes signals in the network downwards based on a probabilistic Winner-Take-All model. Fong and Vedaldi~\cite{fong2017interpretable} and Cao et al.~\cite{cao2015look} learn a perturbation mask that significantly influences the model's output by backpropagating the error signals through the model. Zhou et al.~\cite{zhou2018weakly} extracted fine-detailed class instance activation maps by back-propagating the peak values as top signals to the network downwards in a Winner-Take-All manner. However, the generated maps are less faithful than those produced by CAM-based methods, and such a top-down procedure is complex and computationally expensive.

Sundararajan et al.~\cite{sundararajan2017axiomatic} introduced integrated gradients as a way to quantify a neural network's feature-relevance when making a prediction for a given data point and brought up the concept of missingness in the feature space as a critical interpretability concept. Sturmfels et al.~\cite{sturmfels2020visualizing} later discussed the influence of choosing a baseline input for the integrated gradients. Bau et al.~\cite{bau2017network} introduced network dissection to show that individual units are significantly more interpretable than random linear combinations of units. They consider each unit as a concept detector to further evaluate them for semantic segmentation and quantify the interpretability of CNN latent representations. While these studies proposed solutions to fulfil the sensitivity and implementation invariance axioms, they either required the definition of a baseline input, relied on a threshold derived from the training data set, or limited the solution to a binary segmentation task, all of which failed the sanity checks~\cite{adebayo2018sanity,sixt2020explanations}.

%------------------------------------------------------------------------
\section{ADVISE: {AD}aptive {VI}Sual {E}xplanation}
\label{sec:methodology}

Formally, let $f(I;\theta) = \mathbb{E}[y^{c}|I;\theta]$ represents a CNN that classifies images, and $\theta$ denotes its parameters. For the input image $I \in \mathbb{R}^{H \times W \times 3}$, $y^{c}$ is the score for the predicted class $c$, where $H$ and $W$ denote the height and width of $I$, respectively. Let $A \in \mathbb{R}^{U \times V \times K}$ denotes an activation map in the $f$, where $A^{k}$ represents the $k^{\text{th}}$ feature map in $A$, and $U$, $V$, and $K$ denote the height, width, and the number of units of $f$, respectively. The gradient of the predicted score $y^{c}$ with respect to the spatial location $(i, j)$ in the feature map $A$ can be obtained by $\frac{\partial y^{c}}{\partial A_{i,j}}$.

Although the visualisation methods that calculate cumulative gradients (i.e.,, a linear weighted summation on all feature maps in $A$) preserve implementation invariance, they do not satisfy sensitivity because they assume a stationary rate variation in the gradients. To preserve both the implementation invariance and sensitivity axioms~\cite{sundararajan2017axiomatic}, we propose computing $\phi_{k}(A)$, which assigns an importance score to the $k^{\text{th}}$ unit in the feature map $A$, indicating how much that feature contributes to the network decision. Then we calculate the linear weighted sum of the feature maps in $A$ that have the same importance score. 

Kernel density estimation (KDE) is a conventional non-parametric signal processing approach for estimating the probability density function of data with an unknown underlying distribution~\cite{parzen1962estimation}. Let $(a_{1}, a_{2}, \cdots, a_{n})$ be the value of the independent distributed gradients in the $k^{\text{th}}$ unit of $A$ that were flattened. The gradient values are changed with respect to the input image $I$ and stacked to form a raw density as in Eq.~\ref{eq:01}

\begin{equation}
    \label{eq:01}
    x_{a} = \frac{1}{n}\sum_{i=1}^{n}\delta(a-a_{i}),
\end{equation}
where $n = U \times V$, and $\{a_{i}\}_{i=1}^{n}$ is represented by the Dirac delta function $\delta(a)$. The kernel density estimate is obtained by convolving a kernel $\mathcal{H}_{\omega_{a}}$ with the variable bandwidth $\omega_{a}$ to the raw density $x_{a}$ using Eq.~\ref{eq:02}.

\begin{equation}
    \label{eq:02}
    \widehat{\lambda}_{a}=\int x_{a-s}\mathcal{H}_{\omega_{a}}(s)\;\mathrm{d}s.
\end{equation}
where $\omega_{a}$ is selected as a fixed bandwidth optimised in a local interval, and the integral $\int$ that does not specify bounds refers to $\int_{-\infty}^{\infty}$. The mean integrated squared error (MISE)~\cite{bowman1984alternative} is a well-known goodness-of-fit metric for optimising the estimated density $\widehat{\lambda}_{a}$ to be as close to the unknown underlying density $\lambda_{a}$ as possible. Motivated by~\cite{shimazaki2010kernel}, we introduce the adaptive MISE (AMISE) criterion at gradient $a$ to select an interval length for local optimisation, determine the goodness-of-fit, and regulate the shape of the function $\lambda_{a}$ as in Eq.~\ref{eq:03}.

\begin{equation}
    \label{eq:03}
    \mathrm{AMISE} = \int \mathbb{E}\left ( \widehat{\lambda}_{u} - \lambda_{u} \right )^{2} \rho_{W}^{u-a}\;\mathrm{d}u,
\end{equation}
where $\mathbb{E}$ is the expected $L_{2}$ loss function, $\widehat{\lambda}_{u} = \int x_{u-s}\mathcal{H}_{\omega}(s)\;\mathrm{d}s$ is the estimated density with a fixed bandwidth $\omega$, and $\rho_{W}^{u-a}$ is a weight function that locates the integration of the squared error in a particular interval $W$ centring at $a$. To minimise AMISE, we introduce the adaptive cost function with respect to $a$ by subtracting the irrelevant term for the choice of $\omega$ as in Eq.~\ref{eq:04}.

\begin{equation}
    \label{eq:04}
    C_{n}^{a}(\omega, W) = \mathrm{AMISE} - \int\lambda_{u}^{2}\rho_{W}^{u-a}\;\mathrm{d}u.
\end{equation}

The optimal fixed bandwidth $\omega^{*}$ is obtained as a minimiser of the estimated cost function that is presented in Eq.~\ref{eq:05}:

\begin{multline}
    \label{eq:05}
    \hat{C_{n}^{a}}(\omega, W) = \frac{1}{n^{2}}\sum_{i,j}\psi_{\omega,W}^{a}(a_{i},a_{j}) \\
    - \frac{2}{n^{2}}\sum_{i\neq j}^{n}\mathcal{H}_{\omega}(a_{i}-a_{j})\rho_{W}^{a_{i}-a},
\end{multline}
where $\psi_{\omega,W}^{a}$ is given in Eq.~\ref{eq:06}.

\begin{equation}
    \label{eq:06}
    \psi_{\omega,W}^{a}(a_{i},a_{j}) = \int\mathcal{H}_{\omega}(u-a_{i})\mathcal{H}_{\omega}(u-a_{j})\rho_{W}^{u-a}\;\mathrm{d}u.
\end{equation}

Since the optimal bandwidth $\omega^{*}$ varies with the length of $W$, we select an interval length of $\frac{\omega^{*}}{\gamma}$\footnote{$\frac{\omega^{*}}{\gamma} = n$ is used in our experiment} that scales with the optimal bandwidth. Here, $\gamma$ is a smoothing parameter, with $\gamma <<1$ causing the variable bandwidth to fluctuate slightly, and $\gamma \sim 1$ causing the variable bandwidth to fluctuate significantly. In our experiments, we consider the $[0,1]$ interval and use the Nadaraya-Watson kernel regression~\cite{nadaraya1964estimating} to obtain the variable bandwidth $\omega_{a}^{\gamma}$ using Eq.~\ref{eq:07}

\begin{equation}
    \label{eq:07}
    \omega_{a}^{\gamma} = \left. \int\rho_{W_{s}^{\gamma}}^{a-s}\bar{\omega}_{s}^{\gamma}\;\mathrm{d}s \middle/ \int\rho_{W_{s}^{\gamma}}^{a-s}\;\mathrm{d}s\right..
\end{equation}
where $W_{a}^{\gamma}$ and $\bar{\omega}_{a}^{\gamma}$ represent the interval length and fixed bandwidth at $a$, respectively. Although the variable bandwidth $\omega_{a}^{\gamma}$ is derived from the same data, the use of different $\gamma$ results in varying degrees of smoothness. In this way, the cost function for the variable bandwidth selected with $\gamma$ is obtained using Eq.~\ref{eq:08}.

\begin{equation}
    \label{eq:08}
    \hat{C}_{n}(\gamma) = \int_{0}^{1}\hat{\lambda}_{a}^{2}\;\mathrm{d}a - \frac{2}{n^{2}}\sum_{i\neq j}\mathcal{H}_{\omega_{a_{i}}^{\gamma}}(a_{i}-a_{j}),
\end{equation}
where $\hat{\lambda}_{a} = \int x_{a-s}\mathcal{H}_{\omega_{a}^{\gamma}}(s)\;\mathrm{d}s$ is an estimated rate, with the variable bandwidth $\omega_{a}^{\gamma}$. The integral is calculated numerically with the stiffness constant $\gamma^{*} = \frac{\sqrt{5}+1}{2}$ that minimises Eq.~\ref{eq:08}. In this study, we use the Gauss density function which is expressed in Eq.~\ref{eq:09}.

\begin{equation}
    \label{eq:09}
    \mathcal{H}_{\omega^{\gamma}}(s) = \frac{1}{\sqrt{2\pi\omega^{\gamma}}}\exp \left ( -\frac{s^{2}}{2(\omega^{\gamma})^{2}} \right ),
\end{equation}

\begin{figure}[!htp]
  \centering
  \begin{subfigure}{0.49\linewidth}
    \includegraphics[width=1\linewidth]{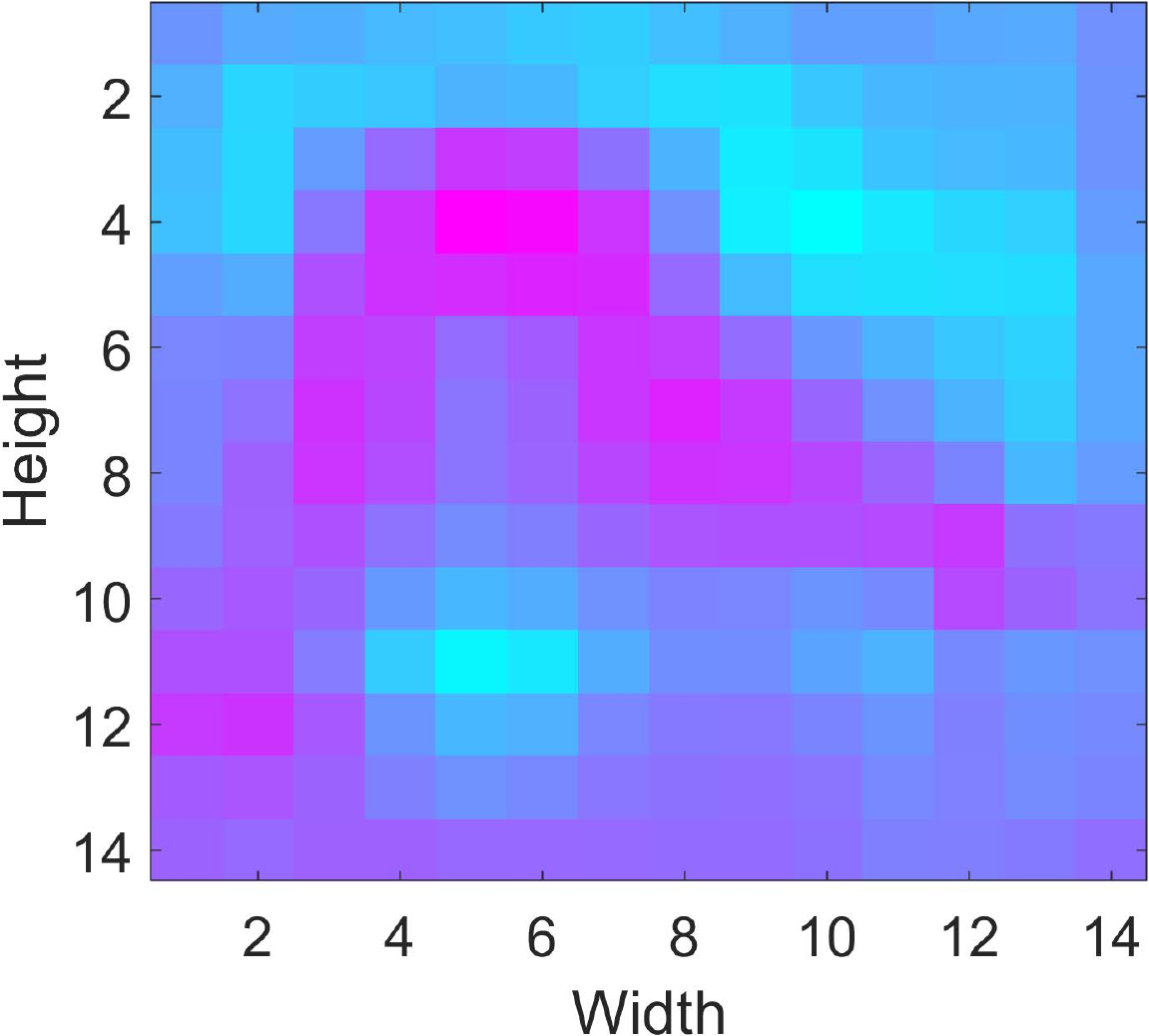}
    \caption{}
    \label{fig:02a}
  \end{subfigure}
  \hfill
  \begin{subfigure}{0.49\linewidth}
    \includegraphics[width=1\linewidth]{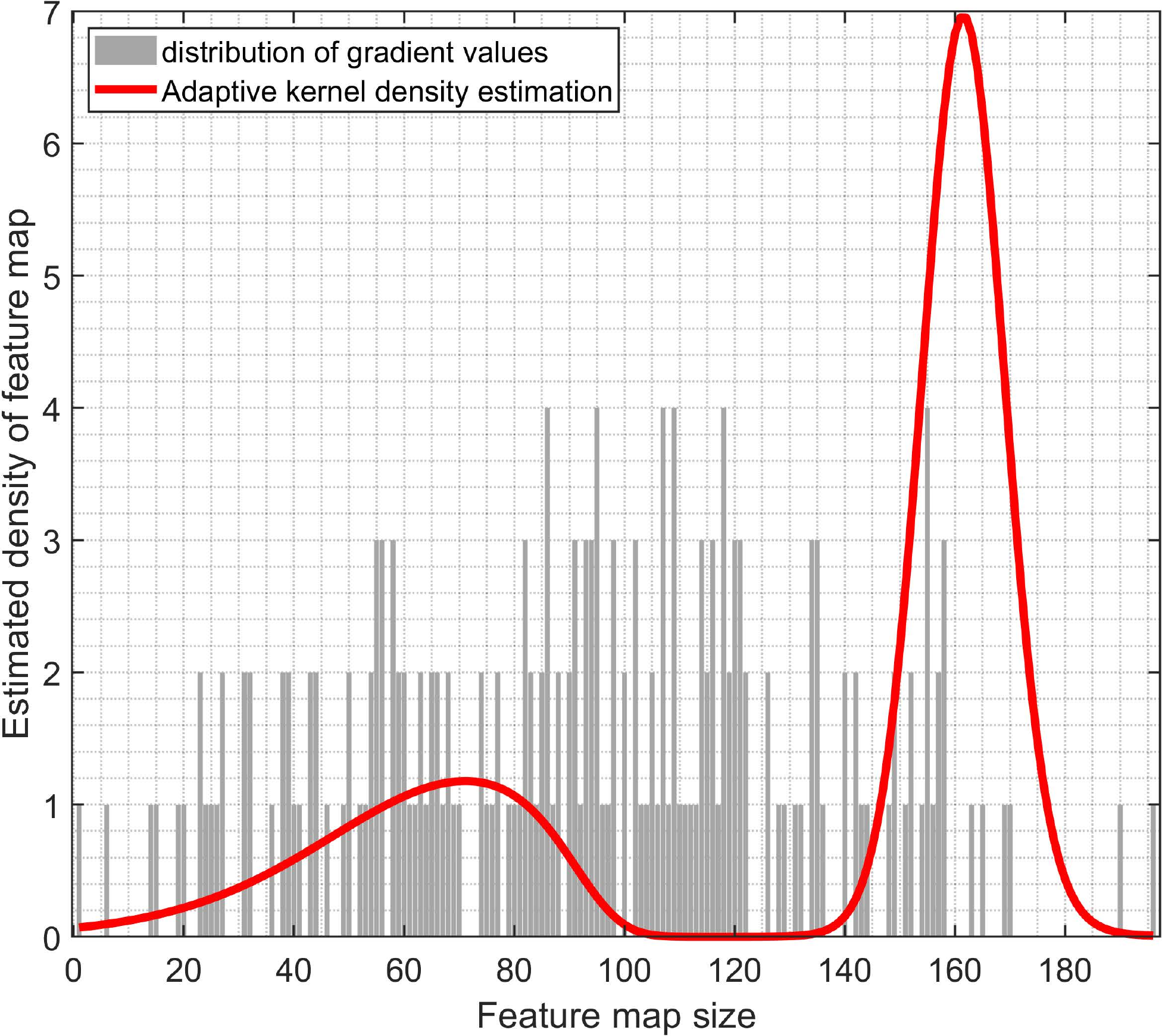}
    \caption{}
    \label{fig:02b}
  \end{subfigure}
  \caption{(a) The $265^{\text{th}}$ unit of the activation map in the last convolution layer of the VGG16 model for the input image in Figure~\ref{fig:01a}, where gradient values are mapped to colours in the `cool' colour map for better visualisation. (b) Estimated kernel density with variable bandwidth (solid red line) using Eq.~\ref{eq:08}. The grey area represents the underlying distribution of gradient values in the $265^{\text{th}}$ unit of the activation map.}
  \label{fig:02}
\end{figure}

Figure~\ref{fig:02a} depicts one of the activation map units in the VGG16 model's final convolution layer for the input image in Figure~\ref{fig:01a}. Figure~\ref{fig:02b} shows the difference between the underlying gradient value distribution (grey area) at that unit and the estimated density of gradient values (solid red line) using the proposed variable bandwidth kernel density estimation.

%%PLACEMENT

The proposed scoring method that assigns an importance score to the $k^{\text{th}}$ unit in the feature map as well as the visualisation approach (ADVISE) are summarised in Algorithm~\ref{alg:01}.

\RestyleAlgo{ruled}
\SetAlgoNoLine
\LinesNumbered
\begin{algorithm}[!htb]
    \SetKwFunction{FP}{findPeaks}\SetKwFunction{FT}{flatten}\SetKwFunction{MN}{min}\SetKwFunction{MX}{max}\SetKwFunction{FN}{find}\SetKwFunction{RS}{resize}
    \SetKwInOut{Input}{Input}\SetKwInOut{Output}{Output}
    \Input{$A^{U \times V \times K}$ -- Feature map, also known as \textit{activation map} in CNNs.\\
    $y^{c}$ -- predicted class.\\
    [row, col] -- size of input image.}
    \Output{$\phi_{k}(A)$ -- Importance score for units in $A$.\\
    ADVISE -- Feature saliency map(s).}
    \BlankLine
    \For{$k\gets 1$ \KwTo $K$}{
        $\{a_{i}\}_{i=1}^{n} \gets$ \FT{$A$}\;
        \tcp{$n = U \times V$.}
        $\phi_{k}(A) =$ \FP{$\int_{0}^{1}\hat{\lambda}_{a}^{2}\;\mathrm{d}a - \frac{2}{n^{2}}\sum_{i\neq j}\mathcal{H}_{\omega_{a_{i}}^{\gamma}}(a_{i}-a_{j})$}\;
    }
    \BlankLine
    $g = \frac{\partial y^{c}}{\partial A}$\;
    \For{$i\gets$ \MN{$\phi_{k}(A)$}  \KwTo \MX{$\phi_{k}(A)$}}{
        $\text{idx} \gets$ \FN{$\phi_{k}(A) == i$}\;
        $\tilde{A}_{i} = A(:,:,\text{idx})$\;
        $\tilde{w}_{i}^{c} = \frac{1}{n}\sum_{U}\sum_{V}g(:,:,\text{idx})$\;
        $\text{map}_{i} = \mathrm{ReLU}\left ( \sum_{j=1}^{
        |\text{idx}|}\tilde{w}_{i,j}^{c} \cdot \tilde{A}_{i,j} \right)$\;
        \tcp{$|\bullet|$ is the cardinality of $\bullet$}
        $\text{ADVISE}_{i} = $ \RS{$\mathrm{map}_{i}$, $\mathrm{[row, col]}$, $\mathrm{bc}$}\;
        \tcp{`bc' is bicubic interpolation}
    }
    \BlankLine
    \Return{$\phi_{k}(A)$, $\mathrm{ADVISE}$}
    \caption{{AD}aptive {VI}Sual {E}xplanation.}
    \label{alg:01}
\end{algorithm}

\begin{figure*}[!htp]
  \centering
  \begin{subfigure}{1\linewidth}
    \includegraphics[width=1\linewidth]{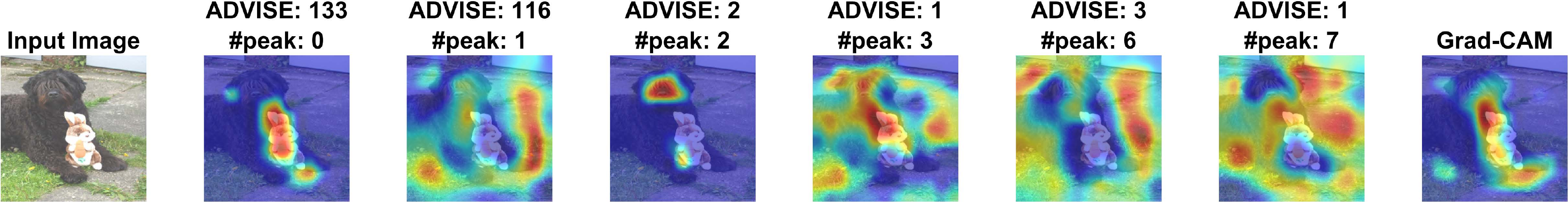}
    \caption{}
    \label{fig:03a}
  \end{subfigure}
  \hfill
  \begin{subfigure}{1\linewidth}
    \includegraphics[width=1\linewidth]{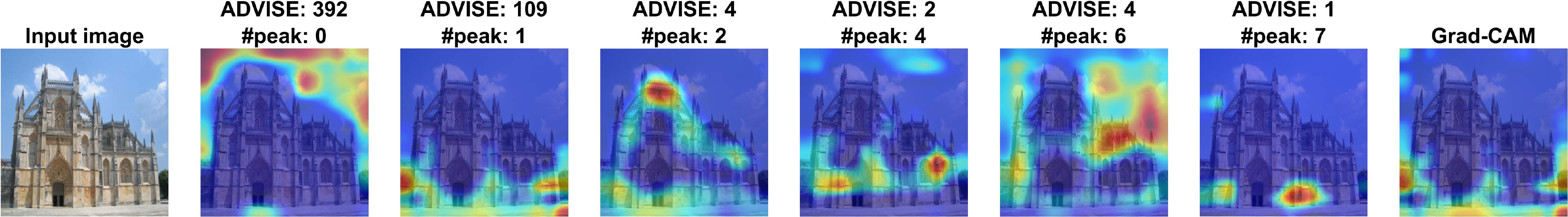}
    \caption{}
    \label{fig:03b}
  \end{subfigure}
  \hfill
  \begin{subfigure}{1\linewidth}
    \includegraphics[width=1\linewidth]{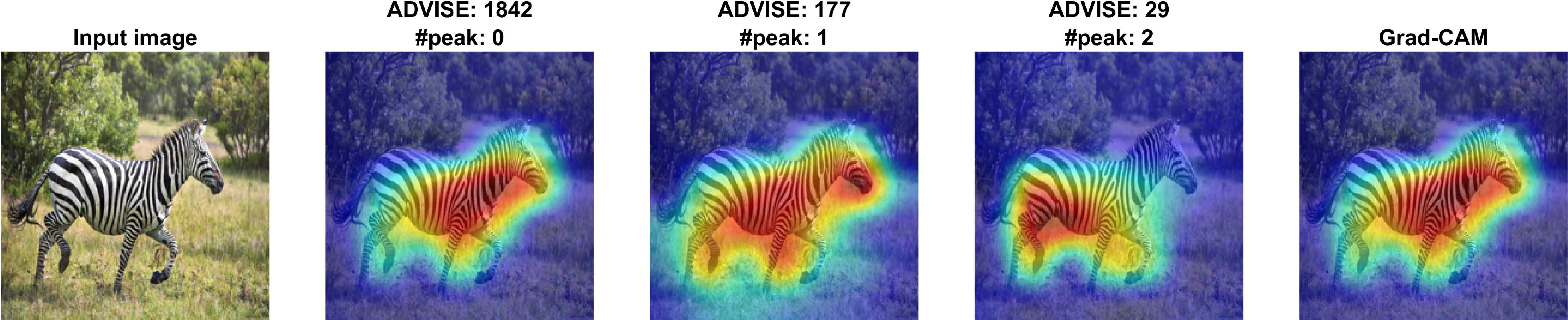}
    \caption{}
    \label{fig:03c}
  \end{subfigure}
  \hfill
  \begin{subfigure}{1\linewidth}
    \includegraphics[width=1\linewidth]{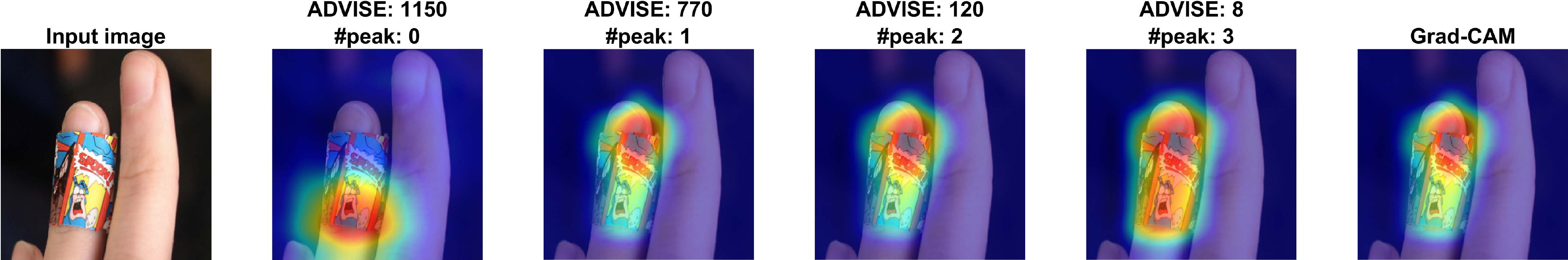}
    \caption{}
    \label{fig:03d}
  \end{subfigure}
  \caption{The outputs of ADVISE and Grad-CAM~\cite{selvaraju2017grad} are compared for four images fed into the pretrained AlexNet~\cite{krizhevsky2012imagenet}, VGG16~\cite{simonyan2015very}, ResNet50~\cite{he2016deep}, and Xception~\cite{chollet2017xception} models on ILSVRC~\cite{russakovsky2015imagenet}. The use of $\phi_k(A)$ on the estimated kernel density and ADVISE show that in the explainability of (a) AlexNet prediction (`Bernese mountain dog'), two units with two peaks work better than Grad-CAM that requires 1000 units, (b) VGG16 prediction (`monastery'), four units with six peaks contribute more than Grad-CAM that requires 512 units, (c) ResNet50 prediction (`Zebra'), 177 units with one peak outperform Grad-CAM, which requires 2048 units, and (d) Xception prediction (`band aid'), eight units with three peaks perform better than Grad-CAM which utilises 2048 units.}
  \label{fig:03}
\end{figure*}

Figure~\ref{fig:03} shows outputs of the proposed method using AlexNet~\cite{krizhevsky2012imagenet}, VGG16~\cite{simonyan2015very}, ResNet50~\cite{he2016deep}, and Xception~\cite{chollet2017xception}, which were trained on the ILSVRC~\cite{russakovsky2015imagenet}.

The results of scoring function $\phi_k(A)$ and the saliency maps generated by ADVISE can highlight three key points. (1) Not all feature map units can contribute equally to the model's prediction, and some of these units may be misleading in some instances (see Figure~\ref{fig:03a}). (2) As Bau et al.~\cite{bau2017network} pointed out, CNNs trained for a specific purpose may encounter the emergence of disentangled representations unrelated to the model's initial intention, complicating interpretation (see Figure~\ref{fig:03b}). As a result, quantifying feature-relevance in conjunction with visualisation can provide adequate answers for users, particularly neural network designers, to underlying questions such as how certain the model is about its decision, if and when it can be trusted, why it makes inevitable mistakes, and in which part of the learning algorithm or parametric space correction should take place. (3) In scenarios such as transfer learning, this mutual explainability approach assists designers in determining which layers should be frozen to achieve better and faster convergence, specifically when the feature map shows less divergence (see Figures~\ref{fig:03c} and~\ref{fig:03d}). In Section~\ref{sec:experiment}, where we introduce quantitative metrics to compare the visualisation approach with the competing ones, we will delve into greater depth on these points. 

%%PLACEMENT
%------------------------------------------------------------------------
\section{Experiments}
\label{sec:experiment}

The proposed method for quantifying feature relevance is applicable to a variety of deep networks. However, we centre our experiments on the image classification task since it allows us to visualise adaptive cumulative gradient attributions and compare ADVISE with attention approaches that focus on global information. ADVISE is tested on a subset of ILSVRC~\cite{russakovsky2015imagenet} with 3,000 images using pretrained AlexNet~\cite{krizhevsky2012imagenet}, VGG16~\cite{simonyan2015very}, ResNet50~\cite{he2016deep}, and Xception~\cite{chollet2017xception} models.

In the absence of ground-truth discriminative features for a trained CNN~\cite{li2021experimental}, objectively identifying which method delivers the best approximation to the usefulness and satisfaction of explanations is still in its early stages. Furthermore, the community has not yet reached a consensus on the impact of explanations on the model's performance, trust, and reliance. A natural assumption is that a well-trained model would make predictions based on the features from the object itself~\cite{arrieta2020explainable}. With this assumption and following quantitative metrics that are used to evaluate image retrieval methods and saliency models, we present a novel evaluation protocol for the visual explanation approaches.

%------------------------------------------------------------------------
\subsection{Evaluation Metrics}
\noindent {\bf (1) Class Sensitivity (CS):} it measures the similarity of saliency maps generated with respect to the top two class scores predicted by the model. We use Pearson’s Correlation Coefficient to measure CS as in Eq.~\ref{eq:10}.

\begin{equation}
    \label{eq:10}
    \mathrm{CS} = \frac{\mathrm{cov}\left ( E(f,I)^{c_{1}},E(f,I)^{c_{2}} \right)}{\sigma\left( E(f,I)^{c_{1}} \right)\times \sigma\left( E(f,I)^{c_{2}} \right)}.
\end{equation}
where $E$, $\mathrm{cov}$, and $\sigma$ denote the explanation map, covariance, and standard deviation, respectively. A good explanation method should have a score near to or below zero, while a score outside the $[-0.5, 0.5]$ range implies that the correlation between two maps is not statistically significant.

\noindent {\bf (2) Hit:} it is a proxy that indicates if the model can retrieve the target class $c$ in its top-5 prediction when it just sees the explanation map and not the entire image. This proxy is formulated in Eq.~\ref{eq:11}.

\begin{equation}
    \label{eq:11}
    \mathrm{Hit} = \left\{ \begin{array}{cl}
    1 & : \ N_{I} \cap M_{I \odot E(f,I)^{c}} \\
    0 & : \ \mathrm{otherwise}
    \end{array} \right.
\end{equation}
where $N_{I}$ is the index of the predicted class $c$ by the model when it just sees the input image as input, and $M_{I \odot E(f,I)^{c}}$ is a set including the top-5 index of the predicted class when the model sees the explanation map. Here, $\odot$ is the Hadamard product.

\noindent {\bf (3) Average Drop (AD):} it measures the average percentage drop in confidence for the target class $c$ when the explanation map $\left (I \odot E(f,I)^{c} \right)$ is fed to the model instead of the input image $I$. This metric is defined in Eq.~\ref{eq:12}, where lower is better.

\begin{equation}
    \label{eq:12}
    \mathrm{AD} = \left. \max\left( 0, (y^{c}-o^{c}) \right) \middle/ y^{c}\right.
\end{equation}
where $o^{c}$ is the predicted score by model to which the the explanation map is fed.

\noindent {\bf (4) Structural similarity index (SSIM):} it is a perception-based measure that considers image degradation as a perceived change in structural information while also considering crucial perceptual phenomena~\cite{wang2004image}. In this context, SSIM measures the structural similarity index between the input image masked by the explanation map and the input image as the reference. This metric returns a value in $(0,1]$, where the higher is better, and is formulated in Eq.~\ref{eq:13}.

\begin{equation}
    \label{eq:13}
    \mathrm{SSIM}(I,\tilde{I}) = \frac{(2\mu_{I}\mu_{\tilde{I}}+e_{1})(2\mathrm{cov}(I,\tilde{I})+e_{2})}{(\mu_{I}^{2}+\mu_{\tilde{I}}^{2}+e_{1})(\sigma_{I}^{2}+\sigma_{\tilde{I}}^{2}+e_{2})}.
\end{equation}
where $\tilde{I} = I \odot E(f,I)^{c}$, and $\mu$ and $\sigma$ are the average and variance, respectively. In order to stabilise the division with weak denominator, $e_{1} = (0.01\cdot L)^{2}$ and $e_{2} = (0.03\cdot L)^{2}$ are used, where $L$ denotes the dynamic range of the pixel values and is set to 255 in this study.

\noindent {\bf (5) Feature similarity index (FSIM):} it uses phase congruency and gradient magnitude, which reflect complementary components of visual image quality, to measure local image quality. This metric also includes a saliency measure for the image gradient feature, which weights each pixel's contribution to the overall quality score. This metric returns a value in $(0,1]$, where the higher is better, and the complete mathematical formulation is given in~\cite{zhang2011fsim}.

\noindent {\bf (6) Mean squared error (MSE):} it is the second error moment and measures the average squared difference between the input image masked by the explanation map and the input image as the reference as in Eq.~\ref{eq:14}.
\begin{equation}
    \label{eq:14}
    \mathrm{MSE}(I,\tilde{I}) = \frac{1}{HW}\sum_{i=1}^{H}\sum_{j=1}^{W}\left( I_{i,j} - \tilde{I}_{i,j} \right)^{2}.
\end{equation}

%%%%%%%%PLACEMENT
\begin{table*}[!htbp]
    \begin{varwidth}[c]{0.63\linewidth}
        \caption{The comparison of the ADVISE with Grad-CAM, Grad-CAM++, Score-CAM, and Layer-CAM visualisation methods on AlexNet, VGG16, ResNet50, and Xception pretrained models on ILSVRC.}
        \label{tbl:01}
        \resizebox{1\textwidth}{!}{
        \begin{tabular}{llcccccccc}
    \hline
    \multirow{2}{*}{\textbf{Architecture}} & \multirow{2}{*}{\textbf{Method}} & \multicolumn{6}{c}{\textbf{Metrics}} & \multicolumn{2}{c}{\textbf{Time (s)}} \\ \cline{3-10} 
     &  & Peak range & AD~\textdownarrow & SSIM~\textuparrow & FSIM~\textuparrow & MSE~\textdownarrow & AVX~\textuparrow & GPU/Parallel & CPU \\ \hline
    \multirow{6}{*}{AlexNet~\cite{krizhevsky2012imagenet}} & \textbf{ADVISE} & 0 -- 8 & \textbf{0.26} & \textbf{0.14} & \textbf{0.38} & \textbf{0.14} & \textbf{0.28} & \textbf{0.69} & 30.3 \\
     & Grad-CAM & N/A & 0.39 & 0.05 & 0.26 & 0.32 & 0.13 & 1.06 & \textbf{1.64} \\
     & Grad-CAM++ & N/A & 0.38 & 0.06 & 0.27 & 0.32 & 0.17 & 1.16 & 2.14 \\
     & Score-CAM & N/A & 0.37 & 0.06 & 0.28 & 0.31 & 0.17 & 1.18 & 2.60 \\
     & Layer-CAM & N/A & 0.33 & 0.07 & 0.31 & 0.28 & 0.19 & 1.48 & 3.33 \\
     & LIME & N/A & 0.39 & 0.05 & 0.26 & 0.32 & 0.13 & 5.71 & 11.85 \\ \hline
    \multirow{6}{*}{VGG16~\cite{simonyan2015very}} & \textbf{ADVISE} & 0 -- 7 & \textbf{0.26} & \textbf{0.14} & \textbf{0.40} & \textbf{0.15} & \textbf{0.29} & \textbf{1.56} & 6.91 \\
     & Grad-CAM & N/A & 0.38 & 0.06 & 0.26 & 0.29 & 0.15 & 1.88 & \textbf{2.66} \\
     & Grad-CAM++ & N/A & 0.38 & 0.07 & 0.27 & 0.28 & 0.19 & 2.01 & 3.36 \\
     & Score-CAM & N/A & 0.37 & 0.09 & 0.30 & 0.29 & 0.22 & 2.21 & 3.87 \\
     & Layer-CAM & N/A & 0.32 & 0.09 & 0.34 & 0.27 & 0.23 & 2.66 & 4.24 \\
     & LIME & N/A & 0.38 & 0.06 & 0.26 & 0.29 & 0.15 & 22.18 & 57.95 \\ \hline
    \multirow{6}{*}{ResNet50~\cite{he2016deep}} & \textbf{ADVISE} & 0 -- 5 & \textbf{0.26} & \textbf{0.15} & \textbf{0.43} & \textbf{0.17} & \textbf{0.31} & \textbf{1.46} & \textbf{6.37} \\
     & Grad-CAM & N/A & 0.33 & 0.10 & 0.34 & 0.24 & 0.23 & 6.22 & 7.77 \\
     & Grad-CAM++ & N/A & 0.36 & 0.11 & 0.35 & 0.24 & 0.26 & 6.62 & 8.56 \\
     & Score-CAM & N/A & 0.35 & 0.11 & 0.37 & 0.22 & 0.27 & 7.02 & 9.18 \\
     & Layer-CAM & N/A & 0.32 & 0.12 & 0.39 & 0.21 & 0.29 & 7.51 & 11.18 \\
     & LIME & N/A & 0.33 & 0.10 & 0.34 & 0.24 & 0.23 & 7.68 & 31.61 \\ \hline
    \multirow{6}{*}{Xception~\cite{chollet2017xception}} & \textbf{ADVISE} & 0 -- 6 & \textbf{0.43} & \textbf{0.12} & \textbf{0.37} & \textbf{0.31} & \textbf{0.24} & \textbf{4.20} & 16.38 \\
     & Grad-CAM & N/A & 0.68 & 0.04 & 0.20 & 0.59 & 0.10 & 5.92 & \textbf{8.12} \\
     & Grad-CAM++ & N/A & 0.65 & 0.04 & 0.21 & 0.59 & 0.11 & 6.03 & 9.10 \\
     & Score-CAM & N/A & 0.64 & 0.05 & 0.21 & 0.57 & 0.13 & 6.56 & 9.70 \\
     & Layer-CAM & N/A & 0.57 & 0.08 & 0.27 & 0.49 & 0.19 & 7.07 & 10.34 \\
     & LIME & N/A & 0.68 & 0.04 & 0.20 & 0.59 & 0.10 & 26.31 & 90.31 \\ \hline
    \end{tabular}%
    }
    \end{varwidth}
    \hfill
    \begin{minipage}[c]{0.35\linewidth}
        \centering
        \captionof{figure}{ADVISE outputs for shallow, middle, and deep layers of (a) VGG16, (b) ResNet50, and (c) Xception pretrained models on ILSVRC.}
        \includegraphics[width=1\linewidth]{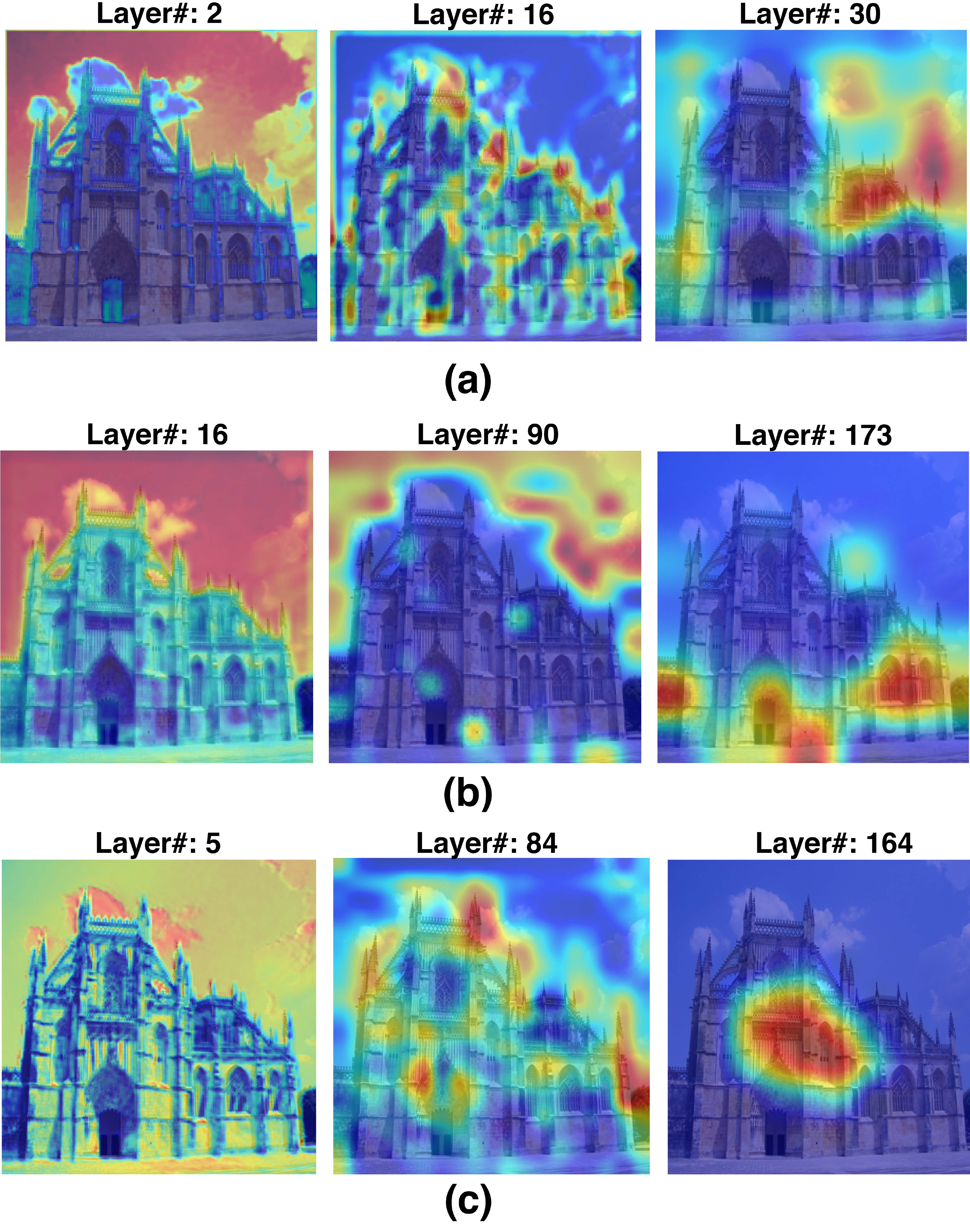}
        \label{fig:04}
        \end{minipage}
\end{table*}

\noindent {\bf (7) AVerage eXplainability (AVX):} it measures the harmonic mean of AD, SSIM, FSIM, and MSE and returns a value in $[0,1]$ to ease of comparison as defined in Eq.~\ref{eq:15}.

\begin{equation}
    \label{eq:15}
    \mathrm{AVX} = 4\left( \frac{1}{1 -\mathrm{AD}} + \frac{1}{\mathrm{SSIM}} + \frac{1}{\mathrm{FSIM}} + \frac{1}{1-\mathrm{MSE}} \right)^{-1}
\end{equation}

Recall that we defined two proxies, $\mathrm{CS}$ and $\mathrm{Hit}$, which allow us to adjust AVX. If $\mathrm{Hit} = 0$ and $\mathrm{CS} \in [-0.5,0.5]$, we define a penalty coefficient $\Delta = 1- |y^{c}-o^{c}|$ and multiply AD, SSIM, FSIM, and MSE by $\Delta$ before measuring the harmonic mean. If $\mathrm{Hit} = 0$ and $\mathrm{CS} \notin [-0.5,0.5]$, we set AD to 1, SSIM to 0, FSIM to 0, and MSE to 1.
%------------------------------------------------------------------------
\subsection{Experimental result}
Table~\ref{tbl:01} shows the comparison of the ADVISE with Grad-CAM~\cite{selvaraju2017grad}, Grad-CAM++~\cite{chattopadhay2018grad}, Score-CAM~\cite{wang2020score}, and Layer-CAM~\cite{jiang2021layercam} visualisation methods on AlexNet~\cite{krizhevsky2012imagenet}, VGG16~\cite{simonyan2015very}, ResNet50~\cite{he2016deep}, and Xception~\cite{chollet2017xception} pretrained models on ILSVRC~\cite{russakovsky2015imagenet}. Despite having a higher performance in classifying ILSVRC than the AlexNet, VGG16, and ResNet50, the Xception model has a lower efficiency in the visual explanation, according to the AVX metric. 

In our quest for this AVX decline in Xception, we examined the saliency maps produced by the ADVISE in shallow, middle, and deep layers (see an example in Figure~\ref{fig:04}). We observed that the saliency maps in the shallow and middle layers highlight low-level visual features distributed across the image, such as edges and blobs. The Xception model, on the other hand, focuses on the centre of a scene in the deep layer, whereas the other models look at different locations. This focus is known as the centre bias in saliency studies~\cite{borji2015reconciling,wolf2021salient}, where most studies revealed that observers prefer to look more often at the centre of the image than at the edges. However, the Xception model's tendency toward centre bias is a double-edged sword. While it is more aligned with human cognitive skills for perceiving visual data, as explained by~\cite{sturmfels2020visualizing}, the centre of mass of the saliency map is the Achilles Heel of many visual explanation methods, with path attribution methods offered to address it~\cite{sundararajan2017axiomatic} but failing the sanity checks~\cite{adebayo2018sanity}.

So what should be done? Although the proposed method and quantitative metrics, which are supported by best practices, can evaluate the performance of different models in visual explanation, we still have a fundamental problem with the lack of \textit{ground-truth explanations}. In fact, we aim to determine which methods best explain our model without knowing how it works. Evaluating supervised models is relatively straightforward since we have a test set. However, evaluating explanations is difficult since we do not exactly know how our model works and do not have the ground-truth for a fair comparison.

%------------------------------------------------------------------------

\subsubsection{Ablation study}
\begin{figure*}[!htp]
  \centering
  \begin{subfigure}{1\linewidth}
    \includegraphics[width=1\linewidth]{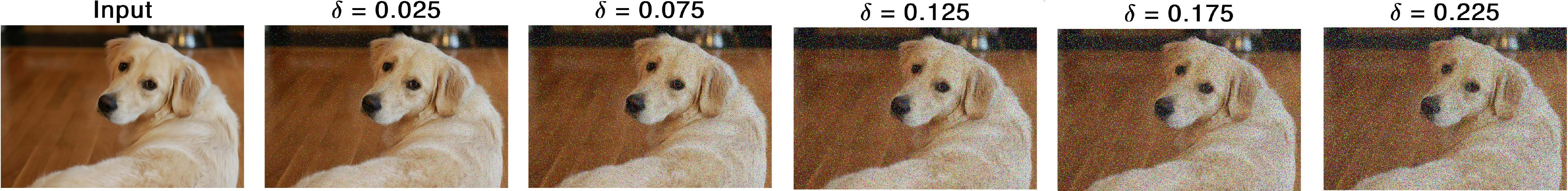}
    \caption{}
    \label{fig:05a}
  \end{subfigure}
  \hfill
  \begin{subfigure}{0.24\linewidth}
    \includegraphics[width=1\linewidth]{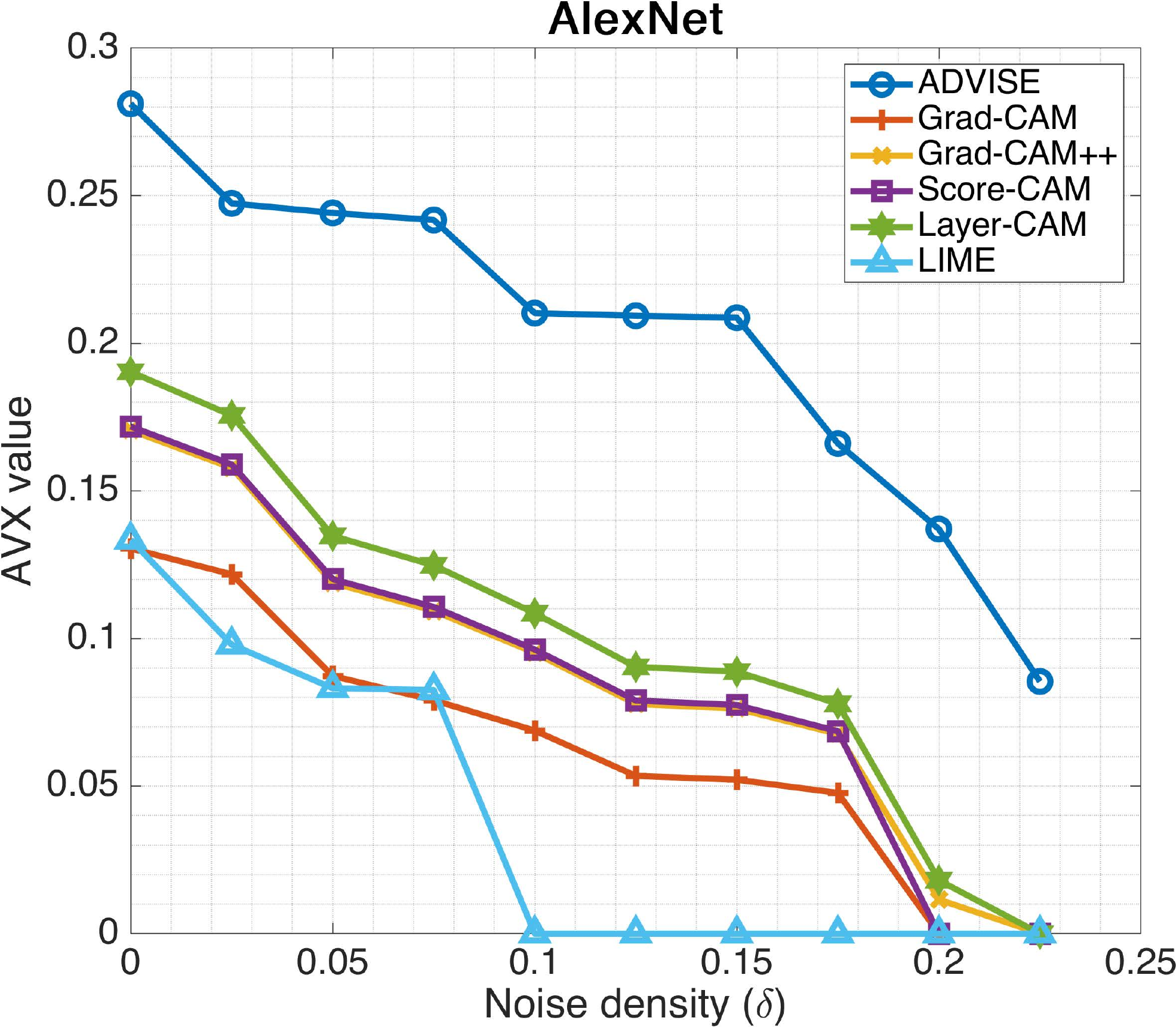}
    \caption{}
    \label{fig:05b}
  \end{subfigure}
  \hfill
  \begin{subfigure}{0.24\linewidth}
    \includegraphics[width=1\linewidth]{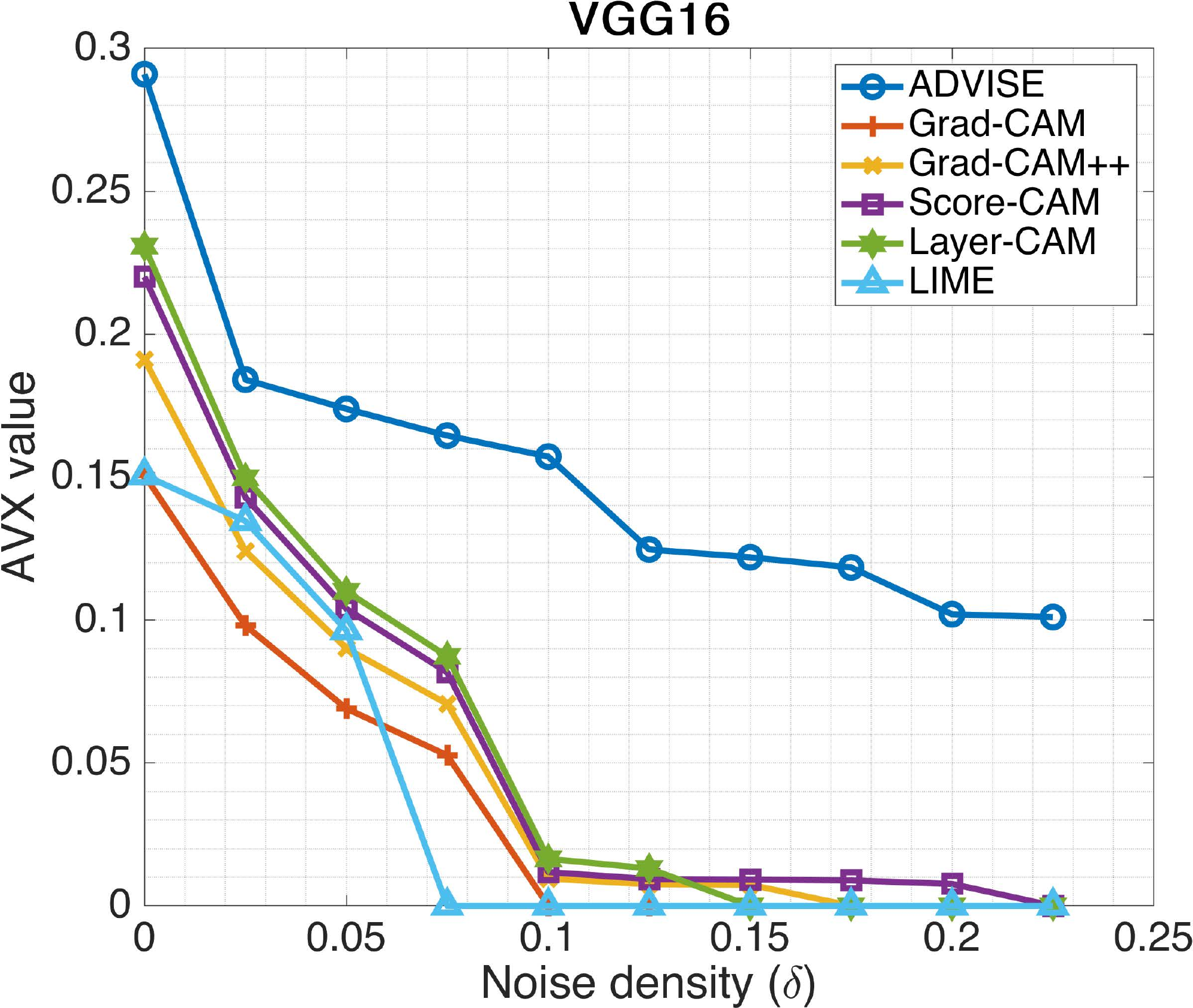}
    \caption{}
    \label{fig:05c}
  \end{subfigure}
  \hfill
  \begin{subfigure}{0.24\linewidth}
    \includegraphics[width=1\linewidth]{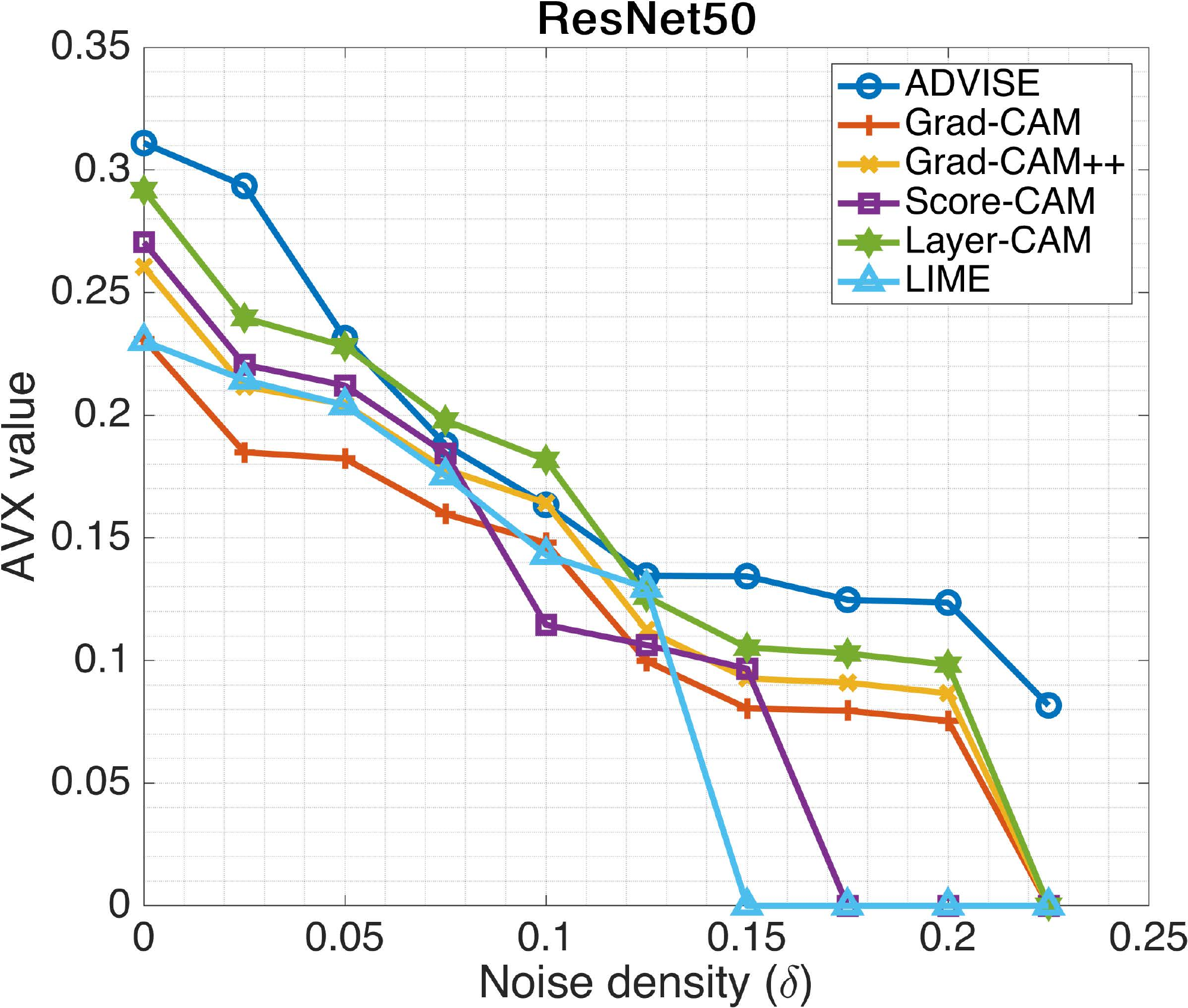}
    \caption{}
    \label{fig:05d}
  \end{subfigure}
  \hfill
  \begin{subfigure}{0.24\linewidth}
    \includegraphics[width=1\linewidth]{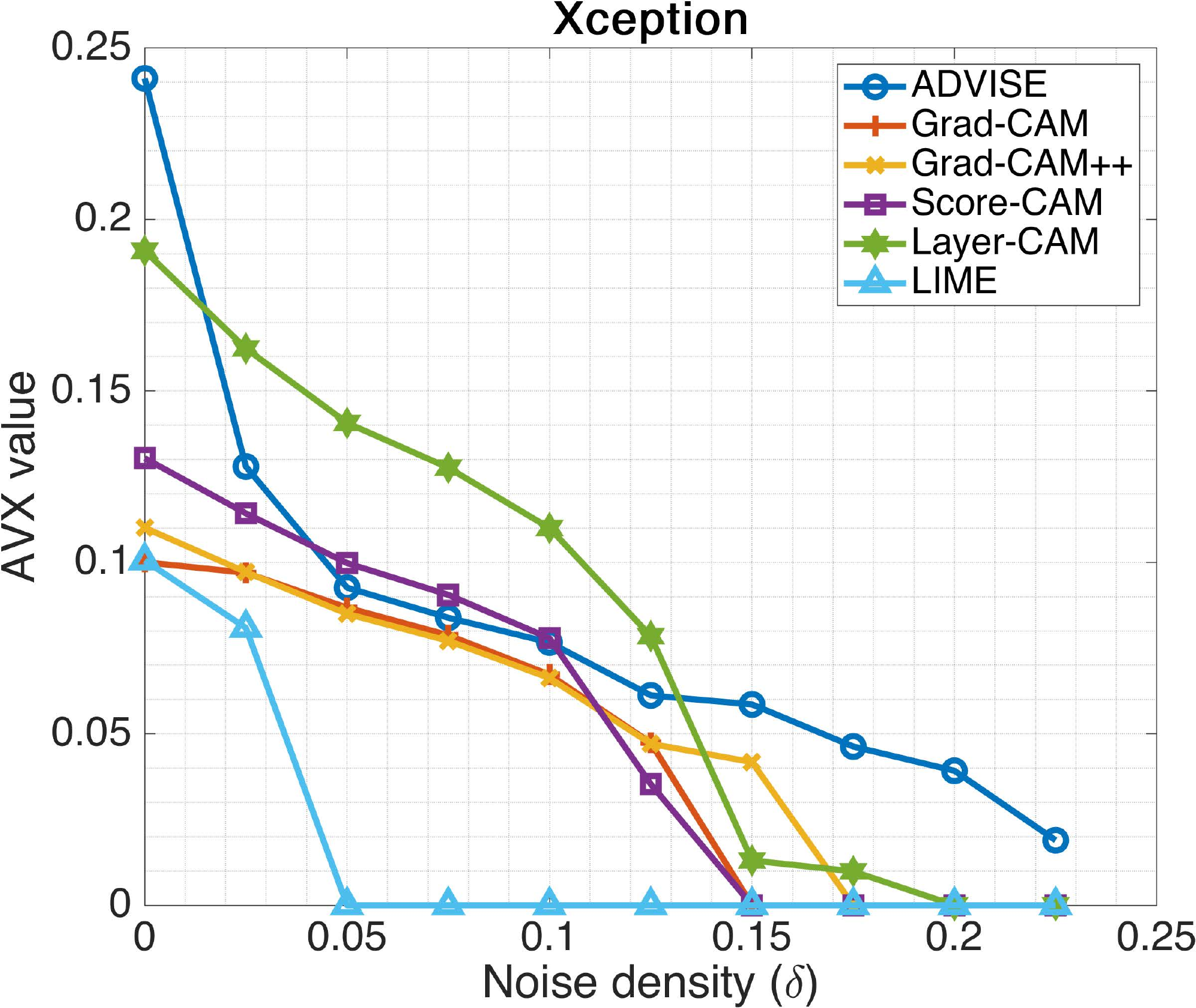}
    \caption{}
    \label{fig:05e}
  \end{subfigure}
  \caption{(a) An ablated image by randomly replacing pixels with the salt and pepper noise with the noise density of $\delta =$~[0.025, 0.075, 0.125, 0.175, 0.225]. (b-e) Changes in the performance of the ADVISE and five additional visual explanation methods in AlexNet, VGG16, ResNet50, and Xception pretrained models on ILSVRC as a function of (AVX, $\delta$).}
  \label{fig:05}
\end{figure*}

The gradient quantifies how much a change in each input dimension affects $f$ prediction in a narrow area around the input. Keeping this in mind, our ablation study is composed of two parts: (1) we ablate the input image by randomly replacing pixels with the salt and pepper noise counterparts; (2) we remove ReLU at the same time to explore the effect of negative gradients on scoring the feature map units and the visual explanation. To do this, all 3,000 images selected from ILSVRC are ablated using the noise density of $\delta = $ [0.025, 0.05, 0.075, 0.1, 0.125, 0.15, 0.175, 0.2, 0.225]. Figure~\ref{fig:05a} depicts an ablated image, and Figure~\ref{fig:05b}--\ref{fig:05e} shows the proposed method's performance compared with other visual explanation methods.

While the AVX value of the ADVISE and other visual explanation methods degrades due to incorporating negative gradients and ablating the input images, the proposed feature scoring method, unlike other methods, could meet the sensitivity axiom~\cite{sundararajan2017axiomatic} in this classification task because the AVX never reached 0. However, we should mention that the pitfall of the ablation test is that if we artificially ablate pixels in an image, we end up with inputs that do not belong to the original data distribution. The question of whether or not users should feed their models with inputs that are not part of the initial training distribution is still being debated~\cite{hooker2019benchmark,sundararajan2020many,janzing2020feature}.
%------------------------------------------------------------------------

%%PLACEMENT

\section{Conclusion}
\label{sec:conclusion}
The significant achievement of Convolutional Neural Networks (CNNs) has resulted in a torrent of computer vision applications. Autonomous systems that can perceive, learn, decide, and act independently are on the horizon for these continuous breakthroughs. However, the incapacity of current approaches to adequately explain their decisions and actions to users limits their effectiveness. Therefore, CNNs must be equipped with the ability to explain their reasoning, characterise their strengths and shortcomings, and convey an understanding of how they will behave in the future. In this study, we have introduced ADVISE, a new explainability method that could quantify and leverage the relevance of each unit of the feature map to provide better visual explanations in CNNs. To this end, we have proposed a method to estimate the kernel density of gradients with an adaptive bandwidth for each unit in the feature map in order to calculate the number of peaks as the unit's relevance score. The cumulative gradient of units with the same relevance score for the class of interest was then calculated to visualise the latent representations in CNNs. We have also proposed a protocol for evaluating the visual explainability of CNN models quantitatively.

In our experiments, we used AlexNet, VGG16, ResNet50, and Xception pretrained on ILSVRC. We have compared ADVISE with the state-of-the-art visual explainable methods and showed that our proposed method outperformed competing approaches in quantifying feature-relevance and visual explainability while maintaining competitive time complexity. Our experiments further demonstrated that ADVISE meets the \textit{sensitivity} and \textit{implementation independence} axioms while passing the sanity checks. 

It is worth mentioning that different metrics have been proposed to evaluate interpretability methods, each with its own set of pros and cons. This lack of consensus on evaluating interpretability methods is related to the fact that we do not know how exactly and transparently our model works and have no specific ground truth against which to compare it. As a result, more experiments on various computer vision tasks and other applications that benefit from the use of deep neural network architectures are required to demonstrate that ADVISE can meet a range of metrics for evaluating interpretability, as we intend to do in the future work.
%-------------------------------------------------------------------------

%%%%%%%%% REFERENCES
{\small
\bibliographystyle{IEEEtran.bst}
\bibliography{reference}
}
\vspace*{-2\baselineskip}
\begin{IEEEbiography}[{\includegraphics[width=1in,height=1.25in,clip,keepaspectratio]{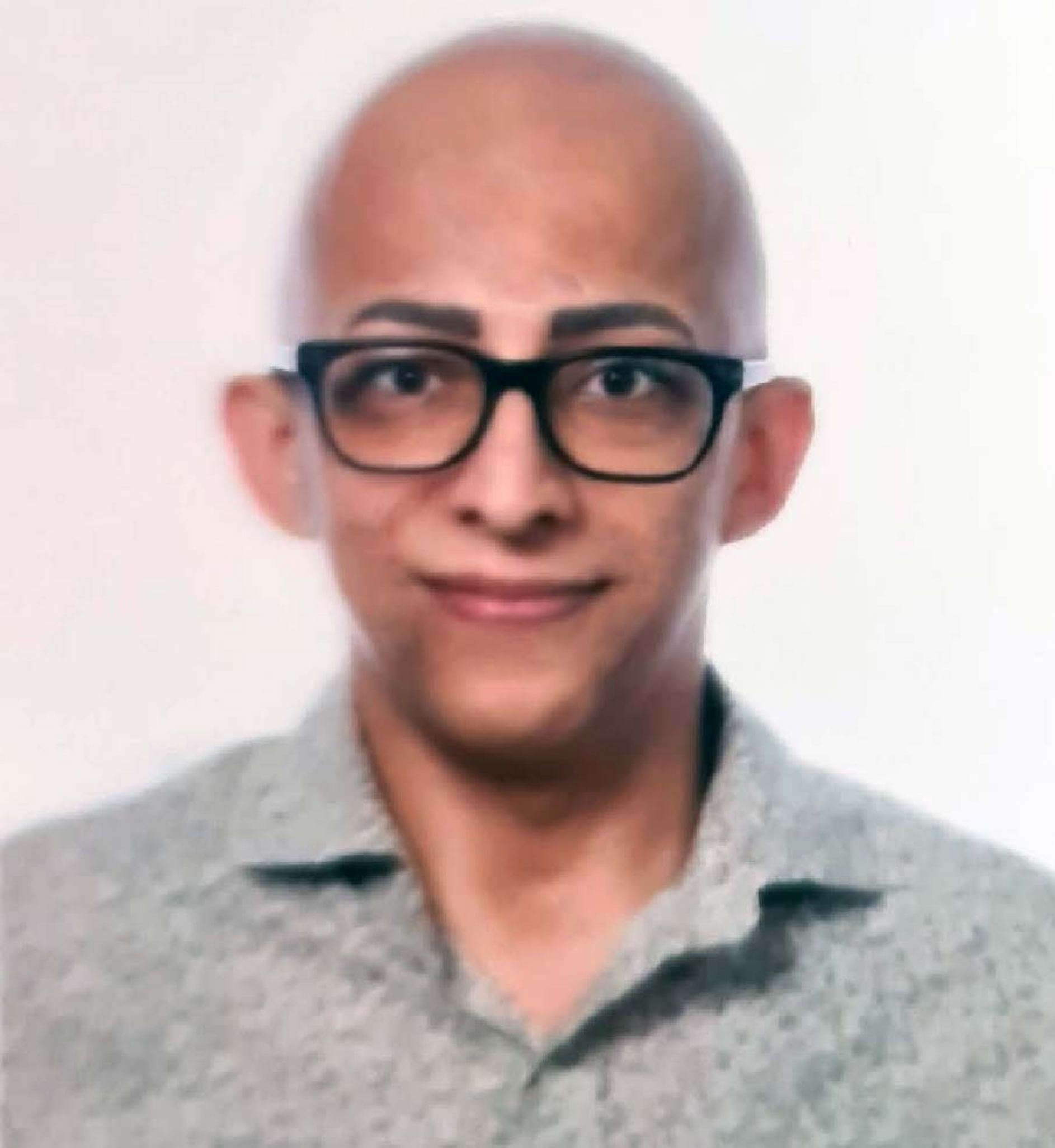}}]{Mohammad Mahdi Dehshibi} received his PhD in Computer Science in 2017 from IAU, Iran. He is currently a research postdoctoral fellow at Universitat Oberta de Catalunya, Spain. He was also a visiting researcher at Unconventional Computing Lab, UWE, Bristol, UK. He has contributed to more than 60 papers published in scientific journals and international conferences. His research interests include Affective Computing, Unconventional Computing, Cellular Automata and Deep Learning.
\end{IEEEbiography}
\vspace*{-2\baselineskip}
\begin{IEEEbiography}[{\includegraphics[width=1in,height=1.25in,clip,keepaspectratio]{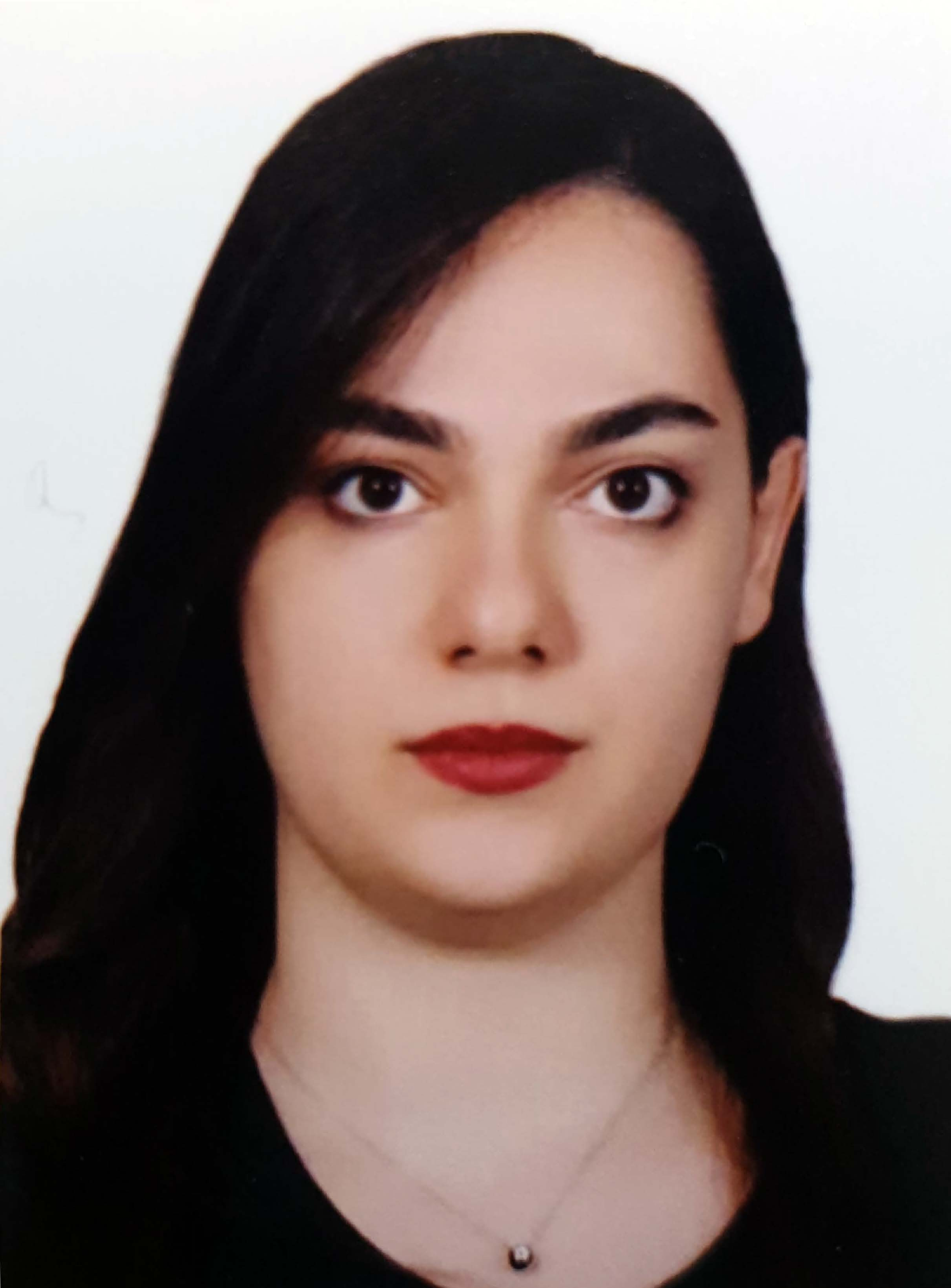}}]{Mona Ashtari-Majlan} received her Master’s degree in Health Systems Engineering from Amirkabir University of Technology, Tehran, in 2021. She is a PhD candidate in computer science at Universitat Oberta de Catalunya, Spain. Her area of interest includes Biomedical Image Processing, Computer Vision, and Deep Learning.
\end{IEEEbiography}
\vspace*{-2\baselineskip}
\begin{IEEEbiography}[{\includegraphics[width=1in,height=1.25in,clip,keepaspectratio]{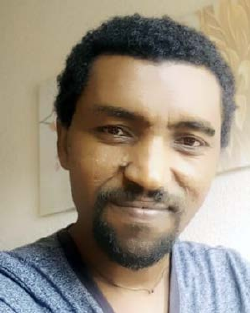}}]{Gereziher Adhane} is currently a PhD student at Universitat Oberta de Catalunya, Spain. He obtained his MSc from Osmania University (India) in 2013/14. His research interests includes deep learning, computer vision and fairness in AI.
\end{IEEEbiography}
\vspace*{-2\baselineskip}
\begin{IEEEbiography}[{\includegraphics[width=1in,height=1.25in,clip,keepaspectratio]{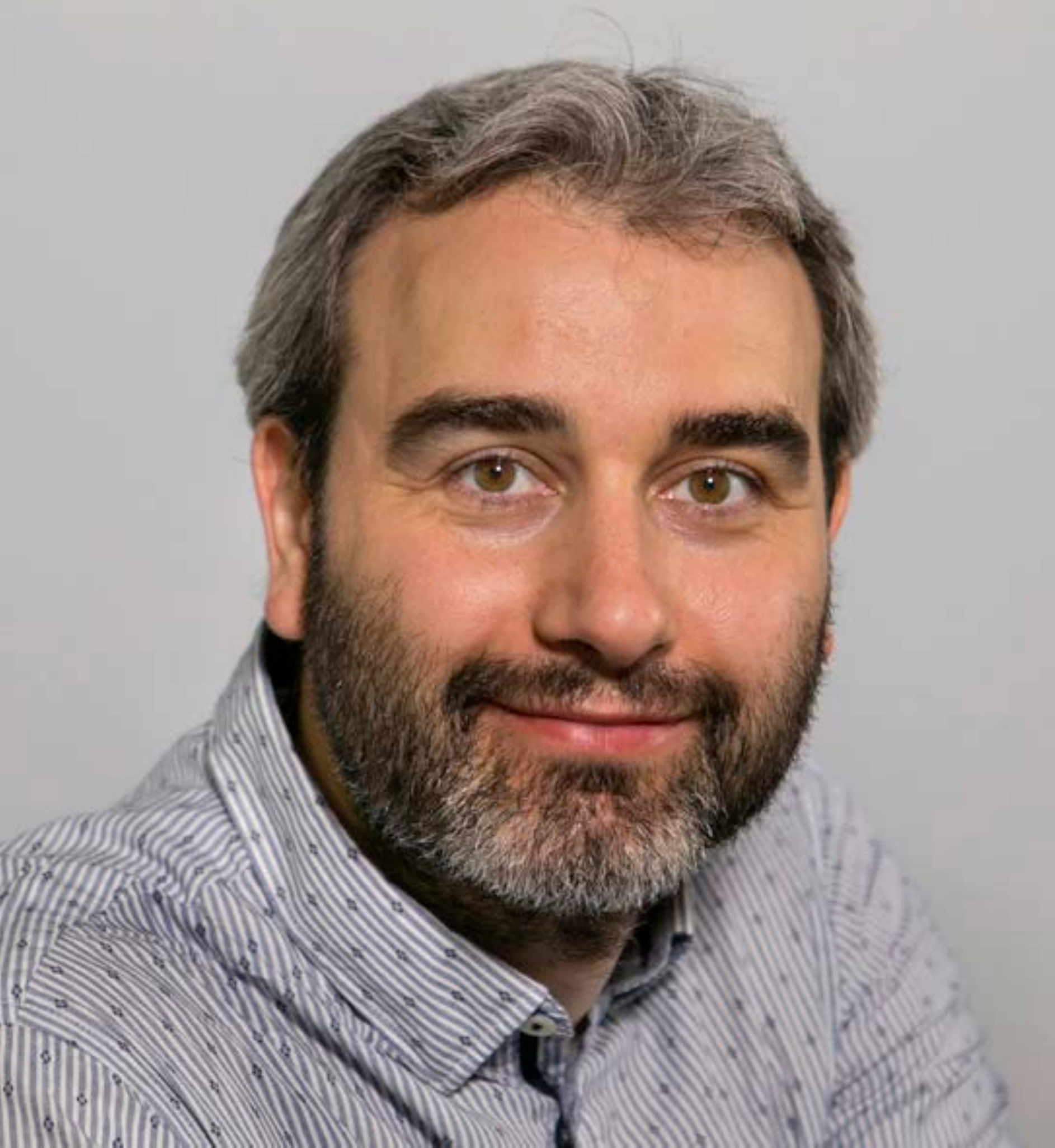}}]{David Masip} is a professor in the Department of Computer Science, Multimedia and Telecommunications at Universitat Oberta de Catalunya (UOC) since February 2007, and the director of the UOC Doctoral School since 2015. He leads the SUNAI (Scene Understanding and Artificial Intelligence) research group. He studied computer science at the Universitat Autonoma de Barcelona (UAB) and received his PhD in September 2005, receiving the UAB's best thesis award in computer science.
\end{IEEEbiography}
\vfill
\end{document}